\documentclass[10pt,twocolumn,letterpaper]{article}

\usepackage[pagenumbers]{cvpr} 

\usepackage{graphicx}
\usepackage{amsmath}
\usepackage{amssymb}
\usepackage{booktabs}
\usepackage{url}            
\usepackage{amsfonts}       
\usepackage{nicefrac}       
\usepackage{microtype}      
\usepackage{xcolor}         
\usepackage{paralist}
\usepackage{caption}
\usepackage{subcaption}
\usepackage{etoolbox}
\usepackage{float}
\usepackage{tabularx}
\usepackage[accsupp]{axessibility}  
\usepackage[pagebackref,breaklinks,colorlinks]{hyperref}

\usepackage[capitalize]{cleveref}
\crefname{section}{Sec.}{Secs.}
\Crefname{section}{Section}{Sections}
\Crefname{table}{Table}{Tables}
\crefname{table}{Tab.}{Tabs.}

\def\cvprPaperID{5938} 
\def\confName{CVPR}
\def\confYear{2023}

\begin{document}

\title{Polynomial Implicit Neural Representations For Large Diverse Datasets}

\author{Rajhans Singh  \qquad Ankita Shukla \qquad Pavan Turaga\\
 \qquad Geometric Media Lab, Arizona State University\\
{\tt\small \{rsingh70, ashukl20, pavan.turaga\}@asu.edu}
}





\twocolumn[{%
\renewcommand\twocolumn[1][]{#1}%
\maketitle
\begin{center}
    \centering
    \captionsetup{type=figure}
    \includegraphics[width=.99\textwidth]{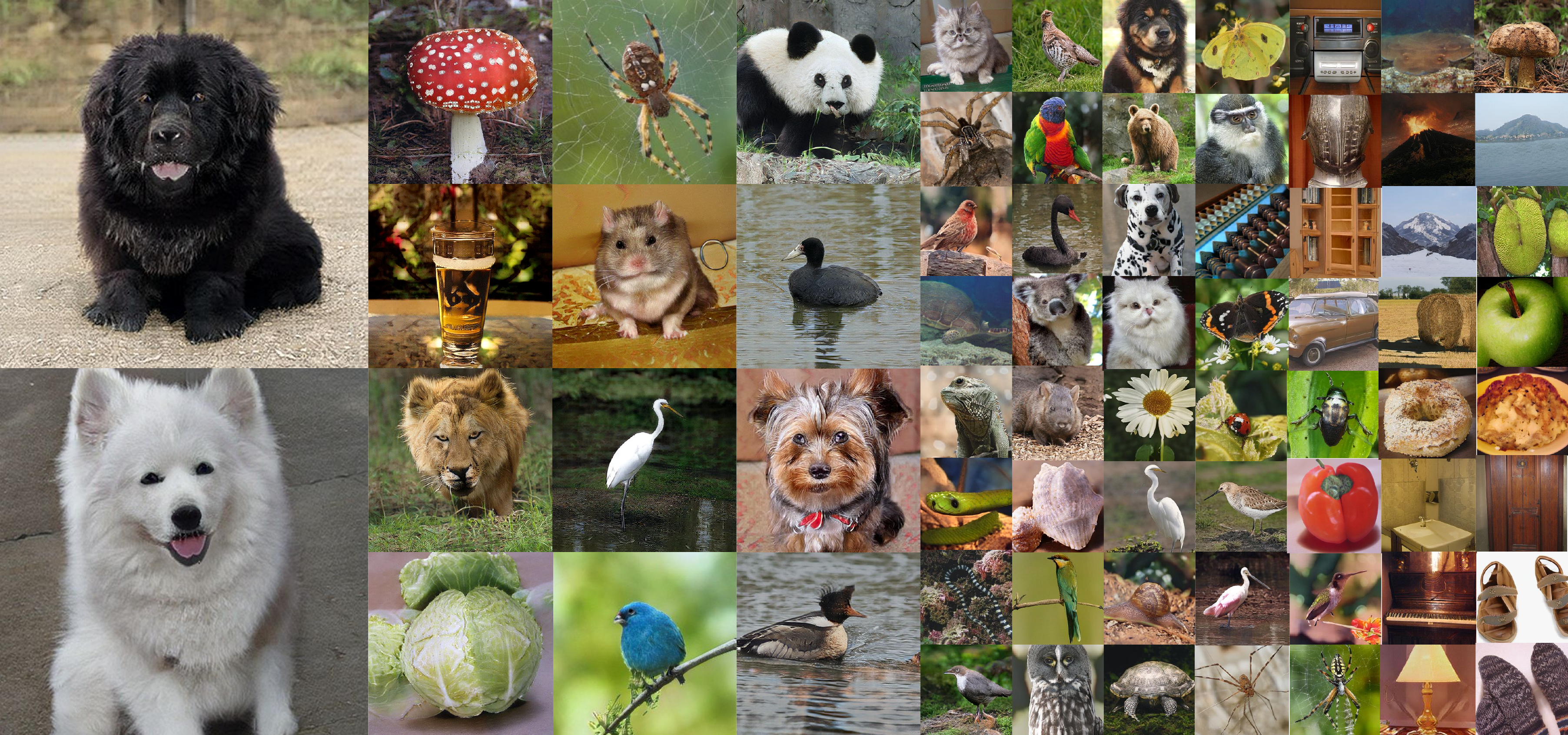}
    \captionof{figure}{Samples generated by our Poly-INR model on the ImageNet dataset at various resolutions. Our model generates images with high fidelity without using convolution, upsample, or self-attention layers, i.e., no interaction between the pixels.}
    \label{fig:teaser}
\end{center}%
}]

\maketitle

\begin{abstract}
\vspace{-3mm}
Implicit neural representations (INR) have gained significant popularity for signal and image representation for many end-tasks, such as superresolution, 3D modeling, and more. Most INR architectures rely on sinusoidal positional encoding, which accounts for high-frequency information in data. However, the finite encoding size restricts the model's representational power. Higher representational power is needed to go from representing a single given image to representing large and diverse datasets. Our approach addresses this gap by representing an image with a polynomial function and eliminates the need for positional encodings. Therefore, to achieve a progressively higher degree of polynomial representation, we use element-wise multiplications between features and affine-transformed coordinate locations after every ReLU layer. The proposed method is evaluated qualitatively and quantitatively on large datasets like ImageNet. The proposed Poly-INR model performs comparably to state-of-the-art generative models without any convolution, normalization, or self-attention layers, and with far fewer trainable parameters. With much fewer training parameters and higher representative power, our approach paves the way for broader adoption of INR models for generative modeling tasks in complex domains. The code is available at \url{https://github.com/Rajhans0/Poly_INR}
\end{abstract}
\vspace{-4mm}
\section{Introduction}


Deep learning-based generative models are a very active area of research with numerous advancements in recent years \cite{kingma2013auto, goodfellow2020generative, dhariwal2021diffusion}. 
Most widely, generative models are based on convolutional architectures. However, recent developments such as implicit neural representations (INR)  \cite{mildenhall2021nerf, sitzmann2020implicit} represent an image as a continuous function of its coordinate locations, where each pixel is synthesized independently. Such a function is approximated by using a deep neural network. INR provides flexibility for easy image transformations and high-resolution up-sampling through the use of a coordinate grid. Thus, INRs have become very effective for $3$D scene reconstruction and rendering from very few training images \cite{mildenhall2021nerf, mescheder2019occupancy, barron2022mip, martin2021nerf,yu2021pixelnerf}. However, they are usually trained to represent a single given scene, signal, or image. Recently, INRs have been implemented as a generative model to generate entire image datasets \cite{anokhin2021image, skorokhodov2021adversarial}. They perform comparably to CNN-based generative models on perfectly curated datasets like human faces \cite{karras2019style}; however, they have yet to be scaled to large, diverse datasets like ImageNet \cite{deng2009imagenet}. 


INR generally consists of a positional encoding module and a multi-layer perceptron model (MLP). The positional encoding in INR is based on sinusoidal functions, often referred to as Fourier features. Several methods \cite{mildenhall2021nerf, sitzmann2020implicit, tancik2020fourier} have shown that using MLP without sinusoidal positional encoding generates blurry outputs, i.e., only preserves low-frequency information. Although, one can remove the positional encoding by replacing the ReLU activation with a periodic or non-periodic activation function in the MLP \cite{sitzmann2020implicit, ramasinghe2022beyond, chng2022gaussian}. However, in INR-based GAN \cite{anokhin2021image}, using a periodic activation function in MLP leads to subpar performance compared to positional encoding with ReLU-based MLP.  

Sitzmann et al. \cite{sitzmann2020implicit} demonstrate that ReLU-based MLP fails to capture the information contained in higher derivatives. This failure to incorporate higher derivative information is due to ReLU's piece-wise linear nature, and second or higher derivatives of ReLU are typically zero. This can be further interpreted in terms of the Taylor series expansion of a given function. The higher derivative information of a function is included in the coefficients of a higher-order polynomial derived from the Taylor series. Hence, the inability to generate high-frequency information is due to the ineffectiveness of the ReLU-based MLP model in approximating higher-order polynomials. 

Sinusoidal positional encoding with MLP has been widely used, but the capacity of such INR can be limiting for two reasons. First, the size of the embedding space is limited; hence only a finite and fixed combination of periodic functions can be used, limiting its application to smaller datasets. Second, such an INR design needs to be mathematically coherent. These INR models can be interpreted as a non-linear combination of periodic functions where periodic functions define the initial part of the network, and the later part is often a ReLU-based non-linear function. Contrary to this, classical transforms (Fourier, sine, or cosine) represent an image by a linear summation of periodic functions. However, using just a linear combination of the positional embedding in a neural network is also limiting, making it difficult to represent large and diverse datasets. Therefore, instead of using periodic functions, this work models an image as a polynomial function of its coordinate location. 

The main advantage of polynomial representation is the easy parameterization of polynomial coefficients with MLP to represent large datasets like ImageNet. However, conventionally MLP can only approximate lower-order polynomials. One can use a polynomial positional embedding of the form $x^py^q$ in the first layer to enable the MLP to approximate higher order. However, such a design is limiting, as a fixed embedding size incorporates only fixed polynomial degrees. In addition, we do not know the importance of each polynomial degree beforehand for a given image. 

Hence, we do not use any positional encoding, but we progressively increase the degree of the polynomial with the depth of MLP. We achieve this by element-wise multiplication between the feature and affine transformed coordinate location, obtained after every ReLU layer. The affine parameters are parameterized by the latent code sampled from a known distribution. This way, our network learns the required polynomial order and represents complex datasets with considerably fewer trainable parameters. In particular,  the key highlights are summarized as follows:
\begin{compactitem}
\item We propose a Poly-INR model based on polynomial functions and design a MLP model to approximate higher-order polynomials.
\item Poly-INR as a generative model performs comparably to the state-of-the-art CNN-based GAN model (StyleGAN-XL \cite{sauer2022stylegan}) on the ImageNet dataset with $~3-4\times$ fewer trainable parameters (depending on output resolution).
\item Poly-INR outperforms the previously proposed INR models on the FFHQ dataset \cite{karras2019style}, using a significantly smaller model.
\item We present various qualitative results demonstrating the benefit of our model for interpolation, inversion, style-mixing, high-resolution sampling, and extrapolation.
\end{compactitem}

\section{Related work}
\noindent\textbf{Implicit neural representations:}
INRs have been widely adopted for $3$D scene representation and synthesis \cite{sitzmann2019scene, mescheder2019occupancy, mildenhall2021nerf}. Following the success of NeRF \cite{mildenhall2021nerf}, there has been a large volume of work on $3$D scene representation from $2$D images \cite{yariv2020multiview, yu2021pixelnerf, sitzmann2021light, martin2021nerf, pumarola2021d, jain2021putting, barron2022mip}. They have also been used for semantic segmentation \cite{fu2022panoptic}, video \cite{park2021nerfies, xian2021space, gao2021dynamic}, audio \cite{gao2021dynamic}, and time-series modeling \cite{fons2022hypertime}. INRs have also been used as a prior for inverse problems \cite{sitzmann2020implicit, reed2021dynamic}. However, most INR approaches either use a sinusoidal positional encoding \cite{mildenhall2021nerf, tancik2020fourier} or a sinusoidal activation function \cite{sitzmann2020implicit}, which limits the model capacity for large dataset representation. In our work, we represent our Poly-INR model as a polynomial function without using any positional encoding.

\begin{figure*}[ht!]
     \centering
         \includegraphics[width=0.95\textwidth]{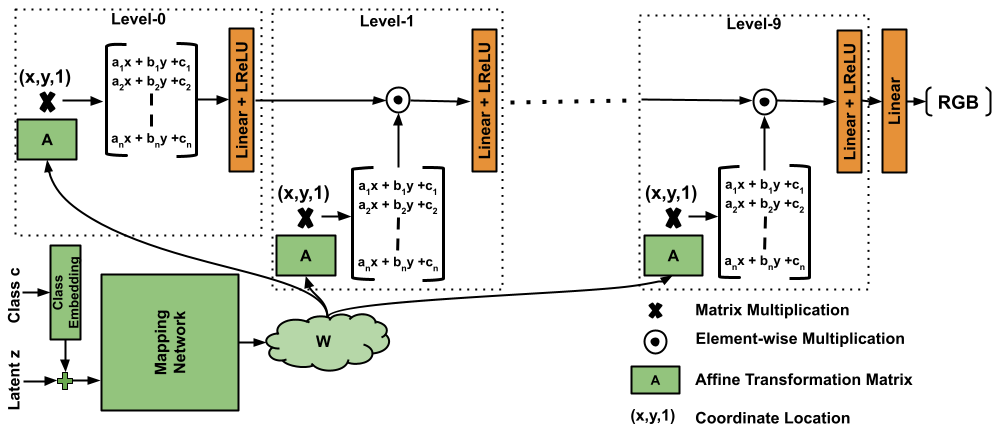}
        \caption{Overview of our proposed Polynomial Implicit Neural Representation (Poly-INR) based generator architecture. Our model consists of two networks: $1)$ Mapping network, which generates the affine parameters from the latent code $z$, and $2)$ Synthesis network, which synthesizes the RGB value for the given pixel location. Our Poly-INR model is defined using only Linear and ReLU layers end-to-end.}
        \label{fig:arch}
\end{figure*}


\noindent\textbf{GANs:} have been widely used for image generation and synthesis tasks \cite{goodfellow2020generative}. In recent work, several improvements have been proposed\cite{radford2015unsupervised, karras2019style,miyato2018spectral,arjovsky2017wasserstein,gulrajani2017improved} over the original architecture. For example, the popularly used StyleGAN \cite{karras2019style} model uses a mapping network to generate style codes which are then used to modulate the weights of the Conv layers. StyleGAN improves image fidelity, as well as enhances inversion \cite{tov2021designing} and image editing capabilities \cite{harkonen2020ganspace}. StyleGAN has been scaled to large datasets like ImageNet \cite{sauer2022stylegan}, using a discriminator which uses projected features from a pre-trained classifier \cite{sauer2021projected}. More recently, transformer-based models have also been used as generators \cite{zhao2021improved, lee2021vitgan}; however, the self-attention mechanism is computationally costly for achieving higher resolution. Unlike these methods, our generator is free of convolution, normalization, and self-attention mechanisms and only uses ReLU and Linear layers to achieve competitive results, but with far fewer parameters.

\noindent\textbf{GANs + coordinates:}
INRs have also been implemented within generative models. For example,  CIPS \cite{anokhin2021image} uses Fourier features and learnable vectors for each spatial location as positional encoding and uses StyleGAN-like weight modulation for layers in the MLP. Similarly, INR-GAN \cite{skorokhodov2021adversarial} proposes a multi-scale generator model where a hyper-network determines the parameters of the MLP. INR-GAN has been further extended to generate an `infinite'-size continuous image using anchors \cite{skorokhodov2021aligning}. However, these INR-based models have only shown promising results on smaller datasets. Our work scales easily to large datasets like ImageNet owing to the significantly fewer parameters. 

Other approaches have combined CNN with coordinate-based features. For example, the Local Implicit Image Function (LIIF) \cite{chen2021learning} and Spherical Local Implicit Image Function (SLIIF) \cite{yoon2022spheresr} use a CNN-based backbone to generate feature vectors corresponding to each coordinate location. Arbitrary-scale image synthesis \cite{ntavelis2022arbitrary} uses a multi-scale convolution-based generator model with scale-aware position embedding to generate scale-consistent images. StyleGAN model, further extended by \cite{karras2021alias} (StyleGAN-3) to use coordinate location-based Fourier features. In addition, StyleGAN-3 uses filter kernels equivariant to the coordinate grid's translation and rotation. However, the rotation equivariant version of the StyleGAN-3 model fails to scale to ImageNet dataset, as reported in \cite{sauer2022stylegan}. Instead of using convolution layers, the Poly-INR only uses linear and ReLU layers.


\noindent\textbf{Relation to classical geometric moment:} Polynomial functions have been explored earlier in the form of geometric moments for image reconstruction \cite{hu1962visual,teague1980image, honarvar2014image, flusser2009moments}. Unlike the Fourier transform, which uses the sinusoidal functions as the basis, the geometric moment method projects the $2$D image on a polynomial basis of the form $x^py^q$ to compute the moment of order $p+q$. The moment matching method \cite{teague1980image} is generally used for image reconstruction from given finite moments. In moment matching, the image is assumed to be a polynomial function, and the coefficients of the polynomial are defined to match the given finite moments. Similar to geometric moments, we also represent images on a polynomial basis; however, our polynomial coefficients are learned end-to-end and defined by a deep neural network. 


\section{Method}
We are interested in a class of functions that represent an image in the form:
\begin{equation}
\label{eq:poly_form}
\begin{aligned}
G(x,y) =  g_{00}+g_{10}x+g_{01}y+...+g_{pq}x^{p}y^{q},
\end{aligned}
\end{equation}

where, $(x,y)$ is the normalized pixel location sampled from a coordinate grid of size $(H\times W)$, while the coefficients of the polynomial $(g_{pq})$ are parameterized by a latent vector $z$ sampled from a known distribution and are independent of the pixel location. Therefore, to form an image, we evaluate the generator $G$ for all pixel locations $(x,y)$ for a given fixed $z$:
\begin{equation}
\label{eq:image_from_G}
\begin{aligned}
I =  \{G(x,y;z)\,|\,(x,y)\in CoordinateGrid(H,W)\},
\end{aligned}
\end{equation}
where, $CoordinateGrid(H,W)=\{(\frac{x}{W-1},\frac{y}{H-1})\,|\, 0\leq x<W, 0\leq y<H \}$. By sampling different latent vectors $z$, we generate different polynomials and represent images over a distribution of real images.

Our goal is to learn the polynomial defined by Eq. \ref{eq:poly_form} using only Linear and ReLU layers. However, the conventional definition of MLP usually takes the coordinate location as input, processed by a few Linear and ReLU layers. This definition of INR can only approximate low-order polynomials and hence only generates low-frequency information. Although, one can use a positional embedding consisting of polynomials of the form $x^py^q$ to approximate a higher-order polynomial. However, this definition of INR is limiting since a fixed-size embedding space can contain only a small combination of polynomial orders. Furthermore, we do not know which polynomial order is essential to generate the image beforehand. Hence, we progressively increase the polynomial order in the network and let it learn the required orders. We implement this by using element-wise multiplication with the affine-transformed coordinate location at different levels, shown in Fig \ref{fig:arch}. Our model consists of two parts: 1) \textbf{Mapping network}, which takes the latent code $z$ and maps it to affine parameters space $\mathbf{W}$, and 2) \textbf{Synthesis network}, which takes the pixel location and generates the corresponding RGB value.

\noindent\textbf{Mapping Network:}
The mapping network takes the latent code $z \in \mathbb{R}^{64}$ and maps it to the space $\mathbf{W} \in \mathbb{R}^{512}$. Our model adopts the mapping network used in \cite{sauer2022stylegan}. It consists of a pre-trained class embedding, which embeds the one hot class label into a $512$ dimension vector and concatenates it with the latent code $z$. Then the mapping network consists of an MLP with two layers, which maps it to the space $\mathbf{W}$. We use this $\mathbf{W}$ to generate affine parameters by using additional linear layers; hence we call $\mathbf{W}$ as affine parameters space.

\noindent\textbf{Synthesis network:}
The synthesis network generates the RGB $(\mathbb{R}^{3})$ value for the given pixel location $(x,y)$. As shown in Fig. \ref{fig:arch}, the synthesis network consists of multiple levels; at each level, it receives the affine transformation parameters from the mapping network and the pixel coordinate location. At \textit{level-$0$}, we affine transform the coordinate grid and feed it to a Linear layer followed by a Leaky-ReLU layer with $negative\_slope=0.2$. At later levels, we do element-wise multiplication between the feature from the previous level and the affine-transformed coordinate grid, and then feed it to Linear and Leaky-ReLU layers. With the element-wise multiplication at each level, the network has the flexibility to increase the order for $x$ or $y$ coordinate position, or not to increase the order by keeping the affine transformation coefficient $a_j=b_j=0$. In our model, we use $10$ levels, which is sufficient to generate large datasets like ImageNet. Mathematically, the synthesis network can be expressed as follows:
\label{eq:synth}
\begin{align}
G_{syn} = \hdots\sigma(W_2((A_2X)\odot \sigma(W_1((A_1X)\odot\\\nonumber \sigma(W_0(A_0X)))))),
\end{align}
where $X\in \mathbb{R}^{3\times HW}$ is the coordinate grid of size $H\times W$ with an additional dimension for the bias, $A_i \in \mathbb{R}^{n\times 3}$ is the affine transformation matrix from the mapping network for \textit{level-i}, $W_i \in \mathbb{R}^{n\times n}$ is the weight of the linear layer at \textit{level-i}, $\sigma$ is the Leaky-ReLU layer and $\odot$ is element-wise multiplication. Here $n$ is the dimension of the feature channel in the synthesis network, which is the same for all levels. For large datasets like ImageNet, we choose the channel dimension $n=1024$, and for smaller datasets like FFHQ, we choose $n=512$. Note that with this definition, our model only uses Linear and ReLU layers end-to-end and synthesizes each pixel independently.

\begin{table*}[ht!]
\caption{Quantitative comparison of Poly-INR method with CNN-based generative models on ImageNet datasets. (d) compares the number of parameters used in all models at various resolutions. The results for existing methods are quoted from the StyleGAN-XL paper.}
\label{table:quantitative}
\parbox{.5\textwidth}{
\centering
\label{table:imagenet128}
\subcaption{ImageNet $128\times 128$} 
\begin{small}
\resizebox{0.49\textwidth}{!}{\begin{tabular}{ccccccc}
\toprule
\multicolumn{1}{c}{\textbf{Model}} & \multicolumn{1}{c}{\textbf{FID $\downarrow$}} & \multicolumn{1}{c}{\textbf{sFID $\downarrow$} }& \multicolumn{1}{c}{\textbf{rFID $\downarrow$}} & \multicolumn{1}{c}{\textbf{IS $\uparrow$}} & \multicolumn{1}{c}{\textbf{Pr $\uparrow$}} & \multicolumn{1}{c}{\textbf{Rec $\uparrow$}}\\

\midrule
BigGAN &  6.02 &7.18 & 6.09 & 145.83 & 0.86 & 0.35\\
CDM &  3.52 &-&-& 128.80&-&-\\
ADM &  5.91 &5.09 &13.29 &93.31 &0.70 &0.65\\
ADM-G  & 2.97 &5.09& 3.80& 141.37& 0.78& 0.59\\
StyleGAN-XL  &1.81 &3.82 &1.82 &200.55& 0.77 &0.55\\
\hline
\textbf{Poly-INR} & 2.08 & 3.93 & 2.76 & 179.64 & 0.70 & 0.45\\
 \bottomrule

\end{tabular}
}
\end{small}
}
\hfill
\parbox{.5\textwidth}{
\centering
\label{table:imagenet256}
\subcaption{ImageNet $256\times 256$} 
\begin{small}
\resizebox{0.49\textwidth}{!}{\begin{tabular}{ccccccc}
\toprule
\multicolumn{1}{c}{\textbf{Model}} & \multicolumn{1}{c}{\textbf{FID $\downarrow$}} & \multicolumn{1}{c}{\textbf{sFID $\downarrow$} }& \multicolumn{1}{c}{\textbf{rFID $\downarrow$}} & \multicolumn{1}{c}{\textbf{IS $\uparrow$}} & \multicolumn{1}{c}{\textbf{Pr $\uparrow$}} & \multicolumn{1}{c}{\textbf{Rec $\uparrow$}}\\

\midrule
BigGAN &  6.95 &7.36 &75.24 &202.65 &0.87 &0.28\\
ADM &  10.94& 6.02& 125.78& 100.98& 0.69& 0.63\\
ADM-G  & 3.94& 6.14 &11.86 &215.84 &0.83 &0.53\\
DiT-XL/2-G& 2.27 & 4.60& -&278.54& 0.83 & 0.57\\
StyleGAN-XL  &2.30 &4.02 &7.06 &265.12 &0.78& 0.53\\
\hline
\textbf{Poly-INR}  & 2.86 & 4.37 & 7.79 & 241.43& 0.71& 0.39\\
 \bottomrule
\end{tabular}}
\end{small}}
\vspace{0.1in}
\vfill
\parbox{.5\textwidth}{
\centering
\label{table:imagenet512}
\subcaption{ImageNet $512\times 512$} 
\begin{small}
\resizebox{0.49\textwidth}{!}{
\begin{tabular}{ccccccc}
\toprule
\multicolumn{1}{c}{\textbf{Model}} & \multicolumn{1}{c}{\textbf{FID $\downarrow$}} & \multicolumn{1}{c}{\textbf{sFID $\downarrow$} }& \multicolumn{1}{c}{\textbf{rFID $\downarrow$}} & \multicolumn{1}{c}{\textbf{IS $\uparrow$}} & \multicolumn{1}{c}{\textbf{Pr $\uparrow$}} & \multicolumn{1}{c}{\textbf{Rec $\uparrow$}}\\
\midrule
BigGAN &  8.43 &8.13 &312.00 &177.90 &0.88 &0.29\\
ADM & 23.24& 10.19& 561.32& 58.06& 0.73& 0.60\\
ADM-G &  3.85 &5.86 &210.83 &221.72 &0.84 &0.53\\
DiT-XL/2-G&3.04 & 5.04& -&240.82& 0.84 & 0.54\\
StyleGAN-XL & 2.41 &4.06 &51.54& 267.75& 0.77& 0.52\\
\hline
\textbf{Poly-INR} &  3.81 & 5.06 & 54.31 &267.44 & 0.70& 0.34\\
 \bottomrule
\end{tabular}}
\end{small}}
\hfill
\parbox{.5\textwidth}{
\centering
\label{table:params}
\subcaption{Number of parameters in millions (M)} 
\begin{small}
\begin{tabular}{ccccc}
\toprule
\multicolumn{1}{c}{\textbf{Model}} & \multicolumn{1}{c}{\boldmath{$64^2$}}& \multicolumn{1}{c}{\boldmath{$128^2$}}&  \multicolumn{1}{c}{\boldmath{$256^2$}} & \multicolumn{1}{c}{\boldmath{$512^2$}}\\

\midrule
BigGAN & - &141.0& 164.3& 164.7\\
ADM  & 296.0& 422.0& 554.0&559.0\\
DiT-XL  & - & - & 675.0&675.0\\
StyleGAN-XL &134.4 & 158.7& 166.3&168.4\\
\hline
\textbf{Poly-INR} &46.0 &46.0& 46.0 &46.0\\
 \bottomrule
\end{tabular}

\end{small}}
\end{table*}

\begin{table}[ht!]
\caption{Quantitative comparison of Poly-INR method with CNN and INR-based generative models on FFHQ dataset at $256\times256$.}
\label{table:ffhq}
\centering
\begin{small}
\begin{tabular}{cccc}
\toprule
\multicolumn{1}{c}{\textbf{Model}} & \multicolumn{1}{c}{\textbf{params (M)}}&  \multicolumn{1}{c}{\textbf{FID $\downarrow$}} &  \multicolumn{1}{c}{\textbf{Inference Time}}\\
& & &\textbf{(sec/img)}\\
\midrule
StyleGAN2 & 30.0 & 3.83 & 0.016\\
StyleGAN-XL  & 67.9 &2.19& 0.047\\
CIPS & 45.9 &4.38& 0.067\\
INR-GAN & 72.4& 4.95& 0.024\\
\hline
\textbf{Poly-INR} & 13.6 & 2.72& 0.054\\
 \bottomrule
\end{tabular}
\end{small}
\end{table}

\noindent\textbf{Relation to StyleGAN:} StyleGANs \cite{karras2019style,karras2020analyzing, karras2021alias} can be seen as a special case of our formulation. By keeping the coefficients ($a_j$,$b_j$) in the affine transformation matrix of $x$ and $y$ coordinate location equal to zero, the bias term $c_j$ would act as a style code. However, our affine transformation adds location bias to the style code, rather than just using the same style code for all locations in StyleGAN models. This location bias makes the model very flexible in applying a style code only to a specific image region, making it more expressive. In addition, our model differs from the StyleGANs in many aspects. First, our method does not use weight modulation/demodulation or normalizing \cite{karras2020analyzing} tricks. Second, our model does not employ low-pass filters or convolutional layers. Finally, we do not inject any spatial noise into our synthesis network. We can also use these tricks to improve the model's performance further. However, our model's definition is straightforward compared to other GAN models.

\section{Experiments}
The effectiveness of our model is evaluated on two datasets: 1) ImageNet \cite{deng2009imagenet} and 2) FFHQ \cite{karras2019style}. The ImageNet dataset consists of $1.2M$ images over $1K$ classes, whereas the FFHQ dataset contains $\sim 70K$ images of curated human faces. All our models have $64$ dimensional latent space sampled from a normal distribution with mean $0$ and standard deviation $1$. The affine parameters space $\mathbf{W}$ of the mapping network is $512$ dimensions, and the synthesis network consists of $10$ levels with feature dimension $n=1024$ for the ImageNet and $n=512$ for FFHQ. We follow the training scheme of the StyleGAN-XL method \cite{sauer2022stylegan} and use a projected discriminator based on the pre-trained classifiers (DeiT \cite{touvron2021training} and EfficientNet \cite{tan2019efficientnet}) with an additional classifier guidance loss \cite{dhariwal2021diffusion}.

We train our model progressively with increasing resolution, i.e., we start by training at low resolution and continue training with higher resolutions as training progresses. Since the computational cost is less at low resolution, the model is trained for large number of iterations, followed by training for high resolution. Since the model is already trained at low resolution, fewer iterations are needed for convergence at high resolution. 
However, unlike StyleGAN-XL, which freezes the previously trained layers and introduces new layers for higher resolution, Poly-INR uses a fixed number of layers and trains all the parameters at every resolution. 

\subsection{Quantitative results}
We compare our model against CNN-based GANs (BigGAN \cite{brock2018large} and StyleGAN-XL \cite{sauer2022stylegan}) and diffusion models (CDM \cite{ho2022cascaded}, ADM, ADM-G \cite{dhariwal2021diffusion}, and DiT-XL \cite{peebles2022scalable}) on the ImageNet dataset. We also report results on the FFHQ dataset for INR-based GANs (CIPS \cite{anokhin2021image} and INR-GAN \cite{skorokhodov2021adversarial}) as they do not train models on ImageNet. 

\noindent \textbf{Quantitative metrics:} We use Inception Score (IS) \cite{salimans2016improved}, Frechet Inception Distance (FID) \cite{heusel2017gans}, Spatial Frechet Inception Distance (sFID) \cite{nash2021generating}, random-FID (rFID) \cite{sauer2022stylegan}, precision (Pr), and recall (Rec) \cite{kynkaanniemi2019improved}. IS (higher the better) quantifies the quality and diversity of the generated samples based on the predicted label distribution by the Inception network but does not compare the distribution of the generated samples with the real distribution. The FID score  (lower the better) overcomes this drawback by measuring the Frechet distance between the generated and real distribution in the Inception feature space. Further, sFID uses higher spatial features from the Inception network to account for the spatial structure of the generated image. Like StyleGAN-XL, we also use the rFID score to ensure that the network is not just optimizing for IS and FID scores. We use the same randomly initialized Inception network provided by \cite{sauer2022stylegan}. In addition, we also compare our model on the precision and recall metric (higher the better) that measures how likely the generated sample is from the real distribution.

Table \ref{table:quantitative} summarizes the results on the ImageNet dataset at different resolutions. The results for existing methods are quoted from the StyleGAN-XL paper. We observe that the performance of the proposed model is third best after DiT-XL and StyleGAN-XL on the FID and IS metrics. The proposed model outperforms the ADM and BigGAN models at all resolutions and performs comparably to the StyleGAN-XL at $128\times128$ and $256\times256$. We also observe that with the increase in image size, the FID score for Poly-INR drops much more than StyleGAN-XL. The FID score drops more because our model does not add any additional layers with the increase in image size. For example, the StyleGAN-XL uses $134.4$M parameters at $64\times 64$ and $168.4$M at $512\times512$, whereas Poly-INR uses only $46.0$M parameters at every resolution, as reported in Table \ref{table:params}(d). The table shows that our model performs comparably to the state-of-the-art CNN-based generative models, even with significantly fewer parameters. On precision metric, the Poly-INR method performs comparably to other methods; however, the recall value is slightly lower compared to StyleGAN-XL and diffusion models at higher resolution. Again, this is due to the small model size, limiting the model's capacity to represent much finer details at a higher resolution. 

We also compare the proposed method with other INR-based GANs: CIPS and INR-GAN on the FFHQ dataset. Table \ref{table:ffhq} shows that the proposed model significantly outperforms these models, even with a small generator model. Interestingly the Poly-INR method outperforms the StyleGAN-2 and performs comparable to StyleGAN-XL, using significantly fewer parameters. Table \ref{table:ffhq} also reports the inference speed of these models on a Nvidia-RTX-6000 GPU. StyleGANs and INR-GAN use a multi-scale architecture, resulting in faster inference. In contrast, CIPS and Poly-INR models perform all computations at the same resolution as the output image, increasing the inference time. 
 \begin{figure}[]
     \centering
         \includegraphics[width=0.95\columnwidth]{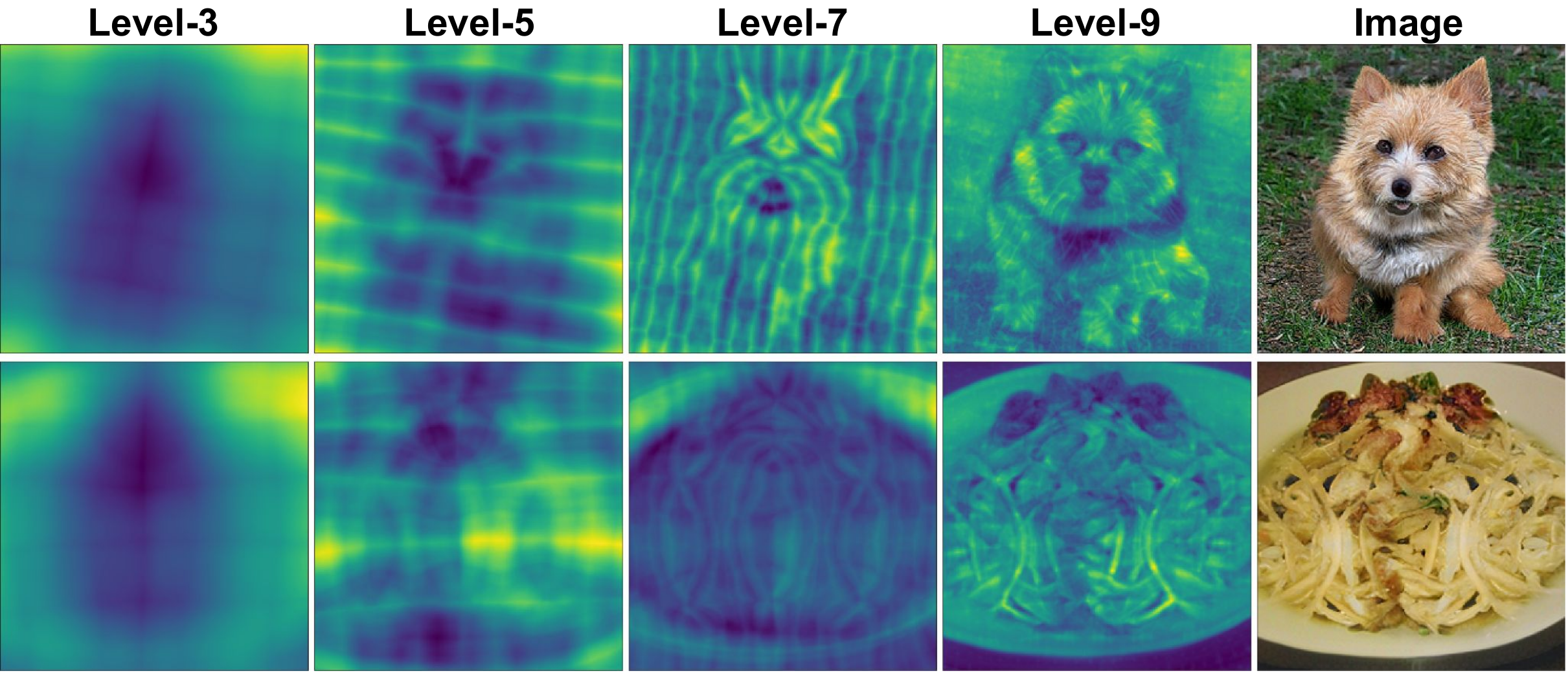}
        \caption{Heat-map visualization at different levels of the synthesis network. At initial levels, the model captures the basic shape of the object, and at higher levels, the image's finer details are captured.}
        \label{fig:heatmap}
\end{figure}

\begin{figure}[]
     \centering
         \includegraphics[width=0.95\columnwidth]{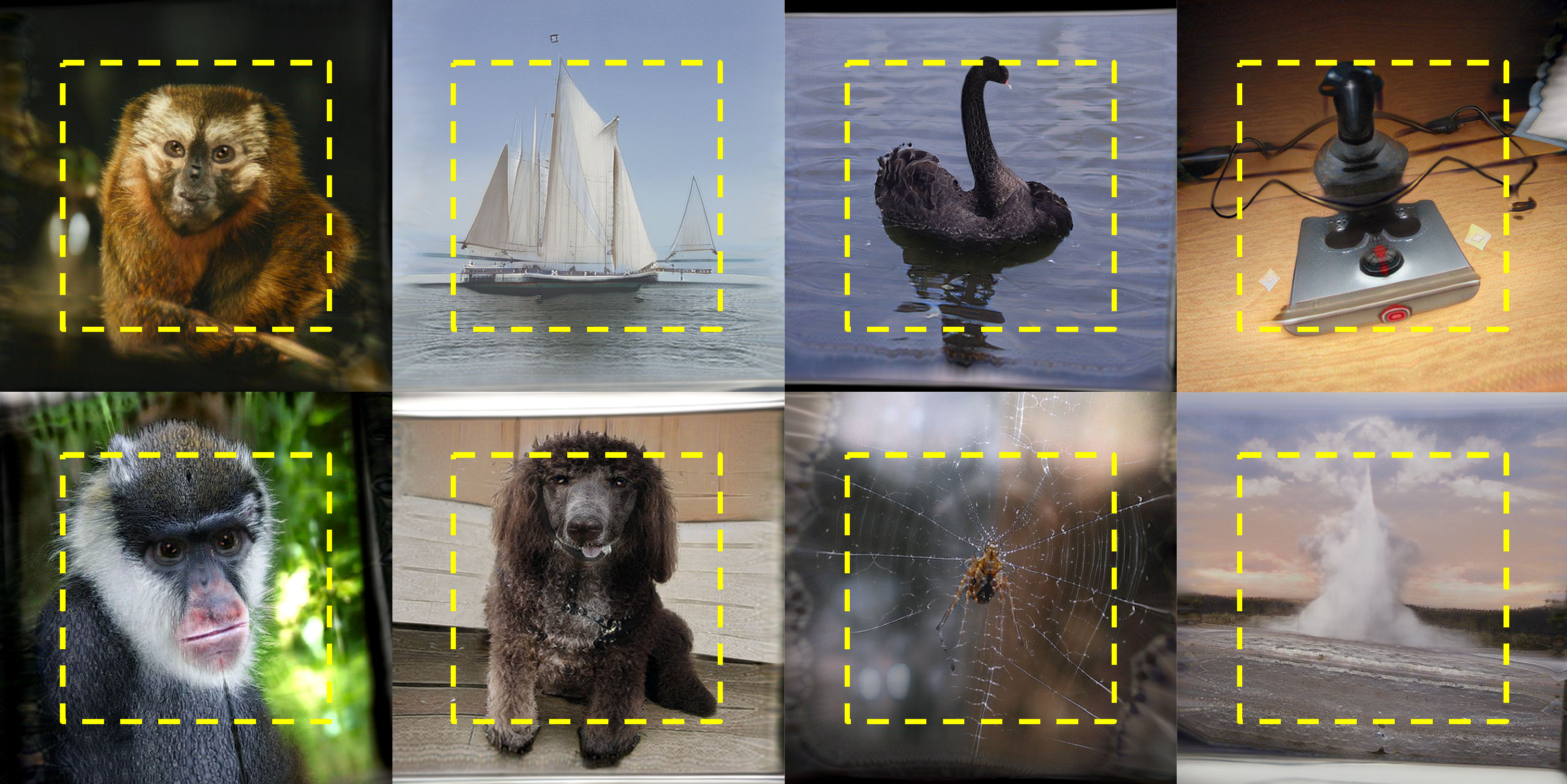}
        \caption{Few example images showing extrapolation outside the image boundary (yellow square). The Poly-INR model is trained to generate images on the coordinate grid $[0,1]^2$. For extrapolation, we use the grid size $[-0.25, 1.25]^2$. Our model generates continuous image outside the conventional boundary.}
        \label{fig:extrapolation}
\end{figure}

\begin{figure*}[]
     \centering
     \vspace{-0.2in}
         \includegraphics[width=0.95\textwidth]{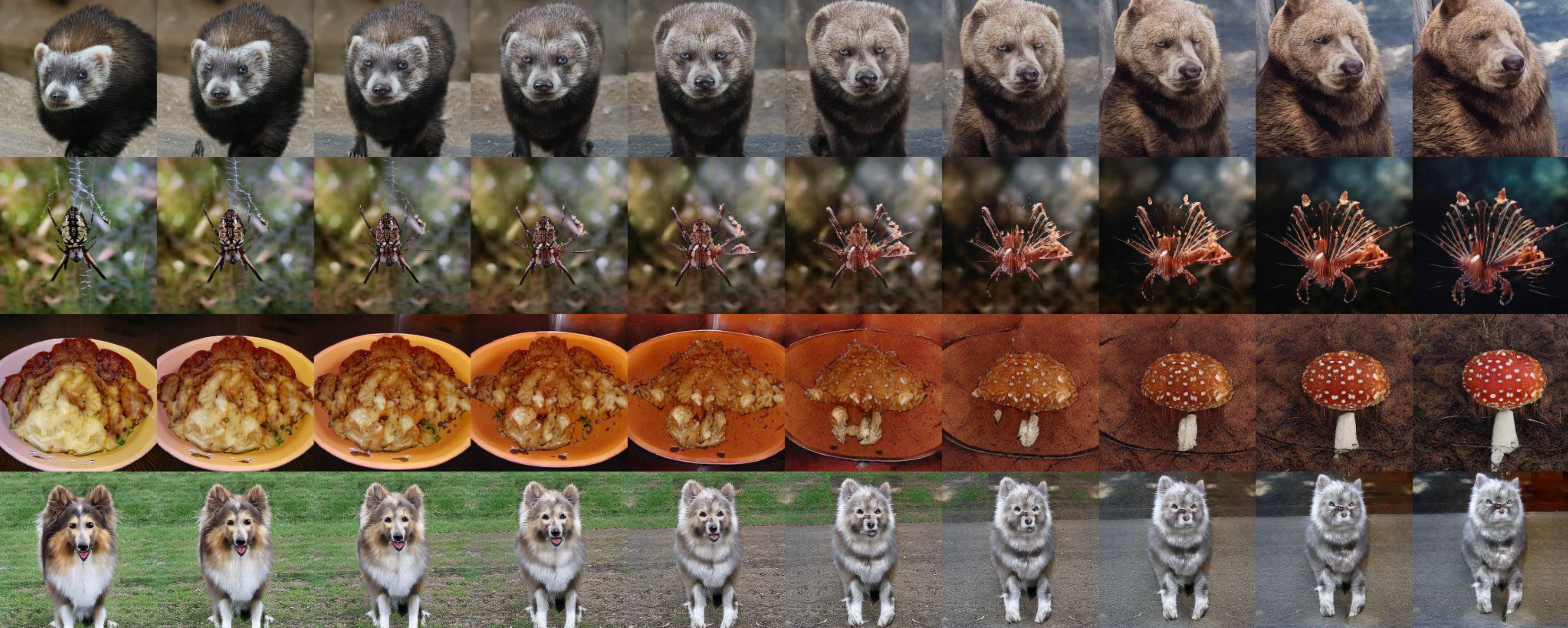}
        \vspace{-0.1in}
        \caption{Linear interpolation between two random points. The first two rows represent interpolation in the latent space, while in the last two, we directly interpolate between the affine parameters. Poly-INR provides smooth interpolation even in a high dimension of affine parameters. Our model generates high-fidelity images similar to state-of-the-art models like StyleGAN-XL but without the need for convolution or self-attention mechanism. Comparisons with existing methods are present in the supplementary material. }
        \label{fig:interpolation}
\end{figure*}

\begin{figure*}[]
     \centering
         \includegraphics[width=0.75\textwidth]{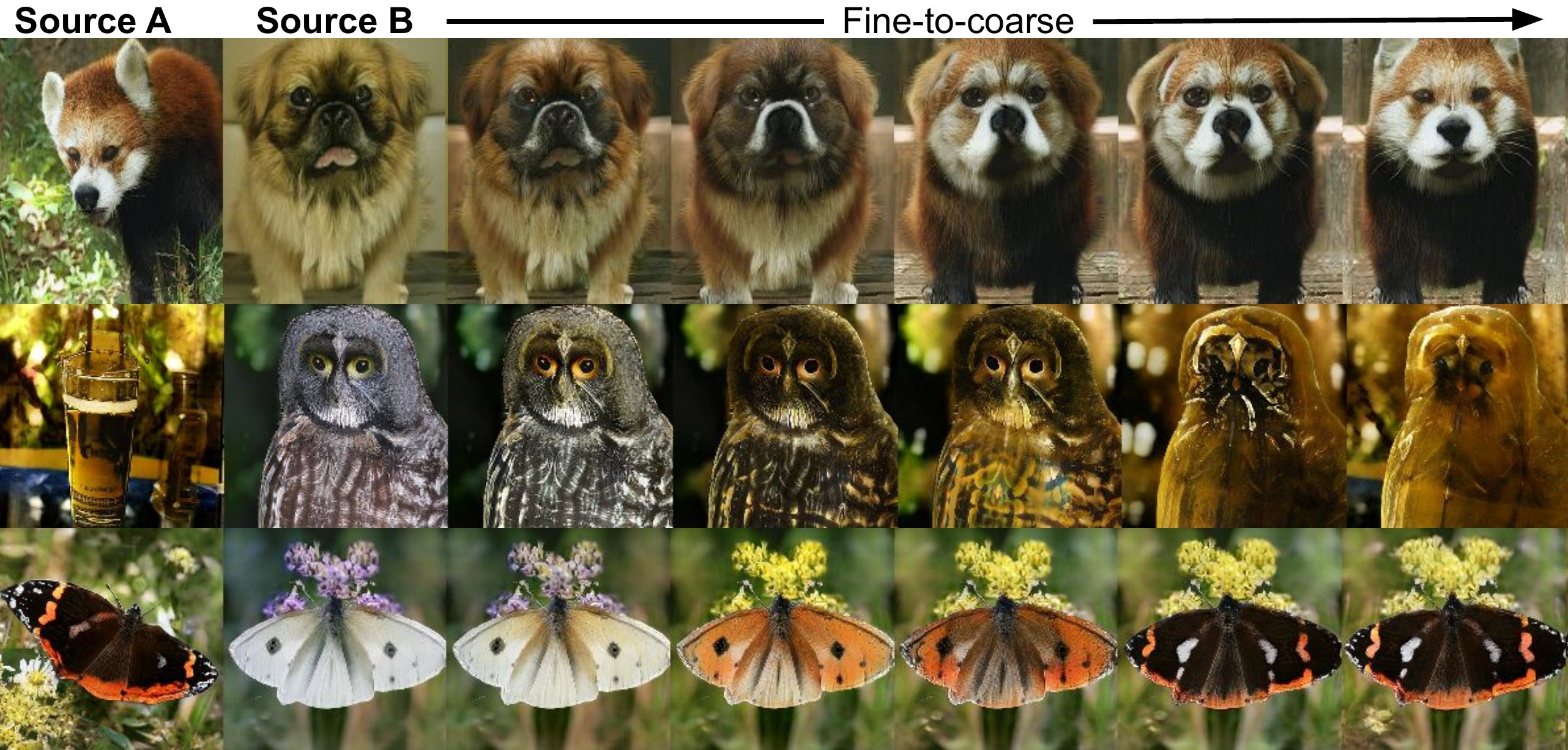}
        \vspace{-0.1in}
        \caption{Source A and B images are generated corresponding to random latent codes, and the rest of the images are generated by copying the affine parameters of source A to source B at different levels. Copying the higher levels' ($8$ and $9$) affine parameters leads to finer style changes, whereas copying the middle levels' ($7$, $6$, and $5$) leads to coarse style changes. }
        \label{fig:stylemixing}
\end{figure*}

\subsection{Qualitative results}
Fig. \ref{fig:teaser} shows images sampled at different resolutions by the Poly-INR model trained on $512 \times 512$. We observe that our model generates diverse images with very high fidelity. Even though the model does not use convolution or self-attention layers, it generates realistic images over datasets like ImageNet. In addition, the model provides flexibility to generate images at different scales by changing the size of the coordinate grid, making the model efficient if low-resolution images are needed for a downstream task. In contrast, CNN-based models generate images only at the training resolution due to the non-equivariant nature of the convolution kernels to image scale. 

\noindent\textbf{Heat-map visualization: }Fig. \ref{fig:heatmap} visualizes the heat-map at different levels of our synthesis network. To visualize a feature as a heat-map, we first compute the mean along the spatial dimension of the feature and use it as a weight to sum the feature along the channel dimension. In the figure, we observe that in the initial levels ($0$-$3$), the model forms the basic structure of the object. Meanwhile, in the middle levels ($4$-$6$), it captures the object's overall shape, and in the higher levels ($7$-$9$), it adds finer details about the object. Furthermore, we can interpret this observation in terms of polynomial order. Initially, it only approximates low-order polynomials and represents only basic shapes. However, at higher levels, it approximates higher-order polynomials representing finer details of the image.

\noindent\textbf{Extrapolation:}
The INR model is a continuous function of the coordinate location; hence we extrapolate the image by feeding the pixel location outside the conventional image boundary. Our Poly-INR model is trained to generate images on the coordinate grid defined by $[0,1]^2$. We feed the grid size $[-0.25, 1.25]^2$ to the synthesis network to generate the extrapolated images. Fig. \ref{fig:extrapolation} shows a few examples of extrapolated images. In the figure, the region within the yellow square represents the conventional coordinate grid $[0,1]^2$. The figure shows that our INR model not only generates a continuous image outside the boundary but also preserves the geometry of the object present within the yellow square. However, in some cases, the model generates a black or white image border, resulting from the image border present in some real images of the training set.

\begin{figure*}[t!]
     \centering
         \includegraphics[width=0.95\textwidth]{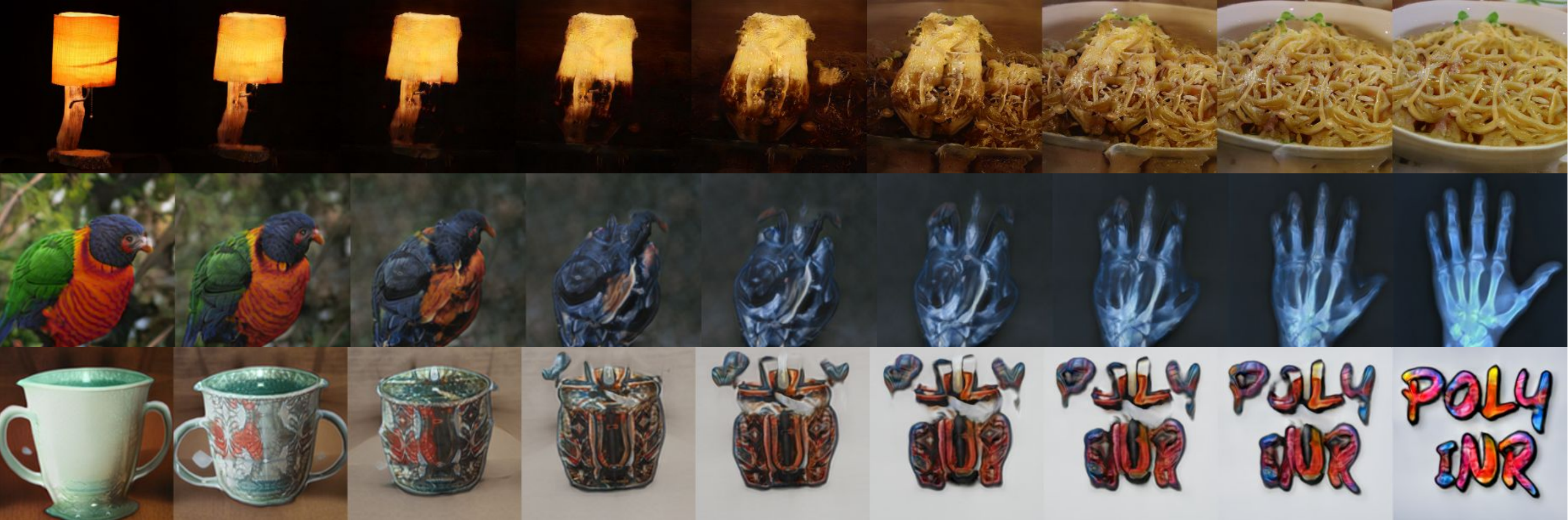}
        \vspace{-0.1in}
        \caption{The Poly-INR model generates smooth interpolation with embedded images in affine parameters space. The leftmost image (first row) is from the ImageNet validation set, and the last two (rightmost) are the OOD images.}
        \label{fig:oodinter}
        \vspace{-0.2in}
\end{figure*}
\begin{figure}[ht!]
     \centering
         \includegraphics[width=0.95\columnwidth]{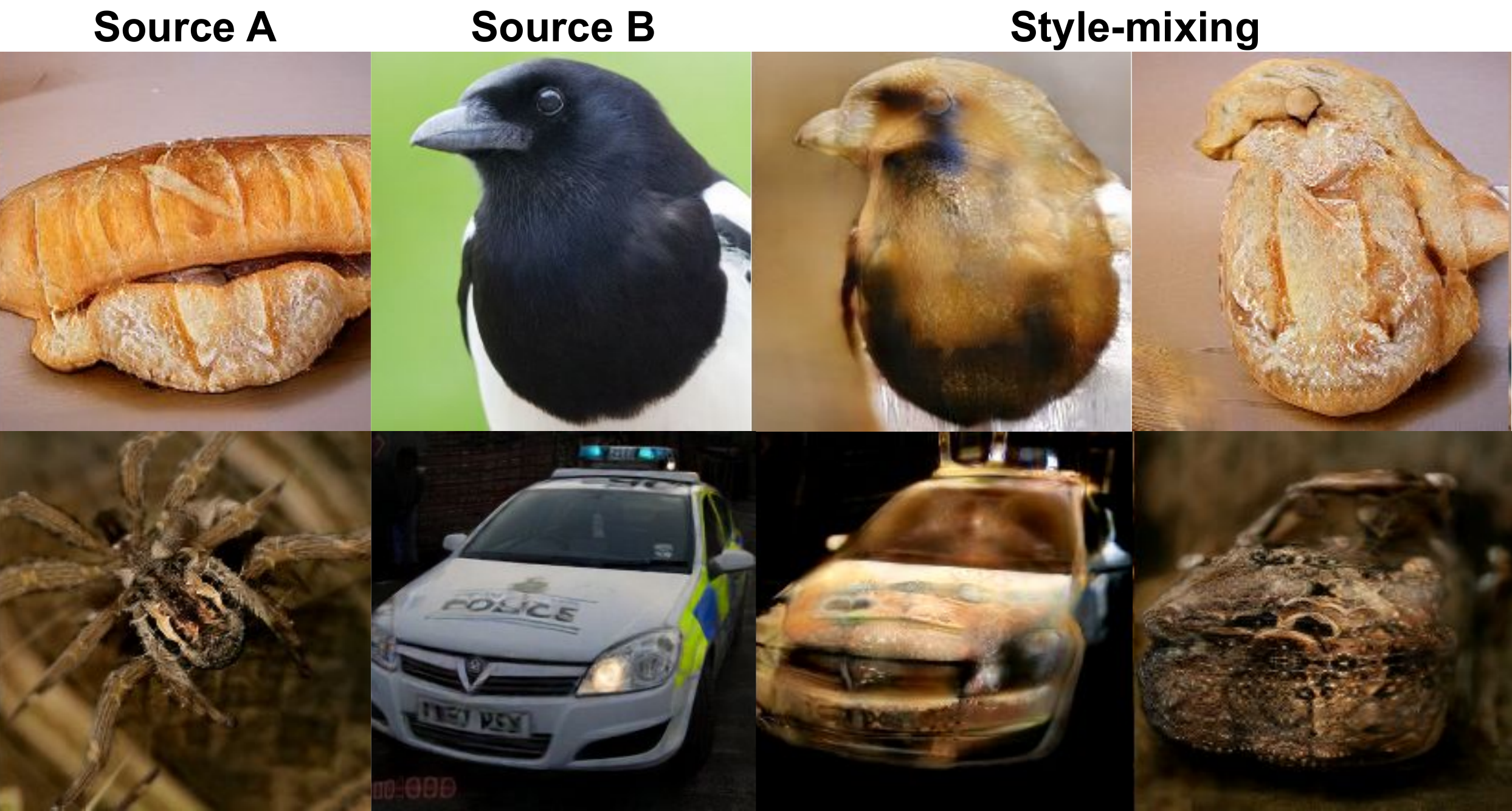}
        \caption{Style-mixing with embedded images in affine parameters space. Source B is the embedded image from the ImageNet validation
set, mixed with the style of randomly sampled source A image.}
        \label{fig:oodstylemix}
\end{figure}
\begin{table}[ht!]
\caption{FID score (lower the better) evaluated at $512 \times 512$ for models trained at a lower resolution and compared against classical interpolation-based upsampling.}
\label{table:highres_sampling}
\centering
\begin{small}
\begin{tabular}{ccccc}
\toprule
\multicolumn{1}{c}{\textbf{Training}} & \multicolumn{1}{c}{\textbf{Nearest}}& 
\multicolumn{1}{c}{\textbf{Bilinear}}&
\multicolumn{1}{c}{\textbf{Bicubic}}&
\multicolumn{1}{c}{\textbf{Poly-INR}}
\\
\textbf{Resolution} & \textbf{Neighbour}& & &\\
\midrule
32$\times$32 & 184.39& 112.28& 73.86 &65.15\\
64$\times$64 & 89.24& 72.41 & 42.97 &36.30\\
 \bottomrule
\end{tabular}
\end{small}
\end{table}

\noindent\textbf{Sampling at higher-resolution:}
Another advantage of using our model is the flexibility to generate images at any resolution, even if the model is trained on a lower resolution. We generate a higher-resolution image by sampling a dense coordinate grid within the $[0,1]^2$ range. Table \ref{table:highres_sampling} shows the FID score evaluated at $512\times512$ for models trained on the lower-resolution ImageNet dataset. We compare the quality of upsampled images generated by our model against the classical interpolation-based upsampling methods. The table shows that our model generates crisper upsampled images, achieving a significantly better FID score than the classical interpolation-based upsampling method. However, we do not observe significant FID score improvement for our Poly-INR model trained on $128\times128$ or higher resolution against the classical interpolation techniques. This could be due to the limitations of the ImageNet dataset, which primarily consists of lower-resolution images than the $512\times512$. We used bilinear interpolation to prepare the training dataset at $512\times512$. As per our knowledge, there are currently no large and diverse datasets like ImageNet with high-resolution images. We believe this performance can be improved when the model has access to higher-resolution images for training. We also compare the upsampling performance with other INR-based GANs by reporting the FID scores at $1024\times1024$ for models trained on FFHQ-$256\times256$ as follows: \textbf{Poly-INR:13.69, INR-GAN: 18.51, CIPS:29.59}. Our Poly-INR model provides better high-resolution sampling than the other two INR-based generators.

\noindent\textbf{Interpolation:}
Fig. \ref{fig:interpolation} shows that our model generates smooth interpolation between two randomly sampled images. In the first two rows of the figure, we interpolate in the latent space, and in the last two rows, we directly interpolate between the affine parameters. In our synthesis network, only the affine parameters depend on the image, and other parameters are fixed for every image. Hence interpolating in affine parameters space means interpolation in INR space. Our model provides smoother interpolation even in the affine parameters space and interpolates with the geometrically coherent movement of different object parts. For example, in the first row, the eyes, nose, and mouth move systematically with the whole face. 

\noindent\textbf{Style-mixing:} Similar to StyleGANs, our Poly-INR model transfers the style of one image to another. Our model generates smooth style mixing even though we do not use any style-mixing regularization during the training. Fig. \ref{fig:stylemixing} shows examples of style-mixing from source A to source B images. For style mixing, we first obtain the affine parameters corresponding to the source A and B images and then copy the affine parameters of A to B at various levels of the synthesis network. Copying affine parameters to higher levels ($8$ and $9$) leads to finer changes in the style, while copying to middle levels ($7$, $6$, and $5$ ) leads to the coarse style change. Mixing the affine parameters at initial levels changes the shape of the generated object. In the figure, we observe that our model provides smooth style mixing while preserving the original shape of the source B object.

\noindent\textbf{Inversion:}
Embedding a given image into the latent space of the GAN is an essential step for image manipulation. In our Poly-INR model, for inversion, we optimize the affine parameters to minimize the reconstruction loss, keeping the synthesis network's parameters fixed. We use VGG feature-based perceptual loss for optimization. We embed the ImageNet validation set in the affine parameters space for the quantitative evaluation. Our Poly-INR method effectively embeds images with high PSNR scores (\textbf{PSNR:}$\mathbf{26.52}$ and \textbf{SSIM:}$\mathbf{0.76}$), better than StyleGAN-XL (PSNR:$13.5$ and SSIM:$0.33$). However, our affine parameters dimension is much larger than the StyleGAN-XL's latent space. Even though the dimension of the affine parameters is much higher, the Poly-INR model provides smooth interpolation for the embedded image. Fig. \ref{fig:oodinter} shows examples of interpolation with embedded images. In the figure, the first row (leftmost) is the embedded image from Val set, and the last two rows (rightmost) are the out-of-distribution images. Surprisingly, our model provides smooth interpolation for OOD images. In addition, Fig. \ref{fig:oodstylemix} shows smooth style-mixing with the embedded images. In some cases, we observe that the fidelity of the interpolated or style-mixed image with the embedded image is slightly less compared to samples from the training distribution. This is due to the large dimension of the embedding space, which sometimes makes the embedded point farther from the training distribution. It is possible to improve interpolation quality further by using the recently proposed pivotal tuning inversion method \cite{roich2022pivotal}, which finetunes the generator's parameters around the embedded point.

\subsection{Discussion}
The proposed Poly-INR model performs comparably to state-of-the-art generative models on large ImageNet datasets without using convolution or self-attention layers. In addition to smooth interpolation and style-mixing, the Poly-INR model provides attractive flexibilities like image extrapolation and high-resolution sampling. In this work, while we use our INR model for $2$D image datasets, it can be extended to other modalities like $3$D datasets.

\noindent\textbf{Challenges:} One of the challenges in our INR method is the higher computation cost compared to the CNN-based generator model for high-resolution image synthesis. The INR method generates each pixel independently; hence all the computation takes place at the same resolution. In contrast, a CNN-based generator uses a multi-scale generation pipeline, making the model computationally efficient. In addition, we observe common GAN artifacts in some generated images. For example, in some cases, it generates multiple heads and limbs, missing limbs, or the object’s geometry is not correctly synthesized. We suspect that the CNN-based discriminator only discriminates based on the object's parts and fails to incorporate the entire shape.




\section{Conclusion}
In this work, we propose polynomial function based implicit neural representations for large image datasets while only using Linear and ReLU layers. Our Poly-INR model captures high-frequency information and performs comparably to the state-of-the-art CNN-based generative models without using convolution, normalization, upsampling, or self-attention layers. The Poly-INR model outperforms previously proposed positional embedding-based INR GAN models. We demonstrate the effectiveness of the proposed model for various tasks like interpolation, style-mixing, extrapolation, high-resolution sampling, and image inversion. Additionally, it would be an exciting avenue for future work to extend our Poly-INR method for $3$D-aware image synthesis on large datasets like ImageNet.   

\section*{Acknowledgements}
This material is based upon work supported by the Defense Advanced Research Projects Agency (DARPA) under Agreement No. HR00112290073. Approved for public release; distribution is unlimited.

{\small
\bibliographystyle{ieee_fullname}
\bibliography{main}

\begin{thebibliography}{10}\itemsep=-1pt

\bibitem{anokhin2021image}
Ivan Anokhin, Kirill Demochkin, Taras Khakhulin, Gleb Sterkin, Victor
  Lempitsky, and Denis Korzhenkov.
\newblock Image generators with conditionally-independent pixel synthesis.
\newblock In {\em Proceedings of the IEEE/CVF Conference on Computer Vision and
  Pattern Recognition}, pages 14278--14287, 2021.

\bibitem{arjovsky2017wasserstein}
Martin Arjovsky, Soumith Chintala, and L{\'e}on Bottou.
\newblock Wasserstein generative adversarial networks.
\newblock In {\em International conference on machine learning}, pages
  214--223. PMLR, 2017.

\bibitem{barron2022mip}
Jonathan~T Barron, Ben Mildenhall, Dor Verbin, Pratul~P Srinivasan, and Peter
  Hedman.
\newblock Mip-nerf 360: Unbounded anti-aliased neural radiance fields.
\newblock In {\em Proceedings of the IEEE/CVF Conference on Computer Vision and
  Pattern Recognition}, pages 5470--5479, 2022.

\bibitem{brock2018large}
Andrew Brock, Jeff Donahue, and Karen Simonyan.
\newblock Large scale gan training for high fidelity natural image synthesis.
\newblock In {\em International Conference on Learning Representations}, 2018.

\bibitem{chen2021learning}
Yinbo Chen, Sifei Liu, and Xiaolong Wang.
\newblock Learning continuous image representation with local implicit image
  function.
\newblock In {\em Proceedings of the IEEE/CVF conference on computer vision and
  pattern recognition}, pages 8628--8638, 2021.

\bibitem{chng2022gaussian}
Shin-Fang Chng, Sameera Ramasinghe, Jamie Sherrah, and Simon Lucey.
\newblock Gaussian activated neural radiance fields for high fidelity
  reconstruction and pose estimation.
\newblock In {\em Computer Vision--ECCV 2022: 17th European Conference, Tel
  Aviv, Israel, October 23--27, 2022, Proceedings, Part XXXIII}, pages
  264--280. Springer, 2022.

\bibitem{deng2009imagenet}
Jia Deng, Wei Dong, Richard Socher, Li-Jia Li, Kai Li, and Li Fei-Fei.
\newblock Imagenet: A large-scale hierarchical image database.
\newblock In {\em 2009 IEEE conference on computer vision and pattern
  recognition}, pages 248--255. Ieee, 2009.

\bibitem{dhariwal2021diffusion}
Prafulla Dhariwal and Alexander Nichol.
\newblock Diffusion models beat gans on image synthesis.
\newblock {\em Advances in Neural Information Processing Systems},
  34:8780--8794, 2021.

\bibitem{flusser2009moments}
Jan Flusser, Barbara Zitova, and Tomas Suk.
\newblock {\em Moments and moment invariants in pattern recognition}.
\newblock John Wiley \& Sons, 2009.

\bibitem{fons2022hypertime}
Elizabeth Fons, Alejandro Sztrajman, Yousef El-Laham, Alexandros Iosifidis, and
  Svitlana Vyetrenko.
\newblock Hypertime: Implicit neural representations for time series.
\newblock In {\em NeurIPS 2022 Workshop on Synthetic Data for Empowering ML
  Research}.

\bibitem{fu2022panoptic}
Xiao Fu, Shangzhan Zhang, Tianrun Chen, Yichong Lu, Lanyun Zhu, Xiaowei Zhou,
  Andreas Geiger, and Yiyi Liao.
\newblock Panoptic nerf: 3d-to-2d label transfer for panoptic urban scene
  segmentation.
\newblock In {\em International Conference on 3D Vision (3DV)}, 2022.

\bibitem{gao2021dynamic}
Chen Gao, Ayush Saraf, Johannes Kopf, and Jia-Bin Huang.
\newblock Dynamic view synthesis from dynamic monocular video.
\newblock In {\em Proceedings of the IEEE/CVF International Conference on
  Computer Vision}, pages 5712--5721, 2021.

\bibitem{goodfellow2020generative}
Ian Goodfellow, Jean Pouget-Abadie, Mehdi Mirza, Bing Xu, David Warde-Farley,
  Sherjil Ozair, Aaron Courville, and Yoshua Bengio.
\newblock Generative adversarial networks.
\newblock {\em Communications of the ACM}, 63(11):139--144, 2020.

\bibitem{gulrajani2017improved}
Ishaan Gulrajani, Faruk Ahmed, Martin Arjovsky, Vincent Dumoulin, and Aaron~C
  Courville.
\newblock Improved training of wasserstein gans.
\newblock {\em Advances in neural information processing systems}, 30, 2017.

\bibitem{harkonen2020ganspace}
Erik H{\"a}rk{\"o}nen, Aaron Hertzmann, Jaakko Lehtinen, and Sylvain Paris.
\newblock Ganspace: Discovering interpretable gan controls.
\newblock {\em Advances in Neural Information Processing Systems},
  33:9841--9850, 2020.

\bibitem{heusel2017gans}
Martin Heusel, Hubert Ramsauer, Thomas Unterthiner, Bernhard Nessler, and Sepp
  Hochreiter.
\newblock Gans trained by a two time-scale update rule converge to a local nash
  equilibrium.
\newblock {\em Advances in neural information processing systems}, 30, 2017.

\bibitem{ho2022cascaded}
Jonathan Ho, Chitwan Saharia, William Chan, David~J Fleet, Mohammad Norouzi,
  and Tim Salimans.
\newblock Cascaded diffusion models for high fidelity image generation.
\newblock {\em J. Mach. Learn. Res.}, 23:47--1, 2022.

\bibitem{honarvar2014image}
Barmak Honarvar, Raveendran Paramesran, and Chern-Loon Lim.
\newblock Image reconstruction from a complete set of geometric and complex
  moments.
\newblock {\em Signal Processing}, 98:224--232, 2014.

\bibitem{hu1962visual}
Ming-Kuei Hu.
\newblock Visual pattern recognition by moment invariants.
\newblock {\em IRE transactions on information theory}, 8(2):179--187, 1962.

\bibitem{jain2021putting}
Ajay Jain, Matthew Tancik, and Pieter Abbeel.
\newblock Putting nerf on a diet: Semantically consistent few-shot view
  synthesis.
\newblock In {\em Proceedings of the IEEE/CVF International Conference on
  Computer Vision}, pages 5885--5894, 2021.

\bibitem{karras2021alias}
Tero Karras, Miika Aittala, Samuli Laine, Erik H{\"a}rk{\"o}nen, Janne
  Hellsten, Jaakko Lehtinen, and Timo Aila.
\newblock Alias-free generative adversarial networks.
\newblock {\em Advances in Neural Information Processing Systems}, 34:852--863,
  2021.

\bibitem{karras2019style}
Tero Karras, Samuli Laine, and Timo Aila.
\newblock A style-based generator architecture for generative adversarial
  networks.
\newblock In {\em Proceedings of the IEEE/CVF conference on computer vision and
  pattern recognition}, pages 4401--4410, 2019.

\bibitem{karras2020analyzing}
Tero Karras, Samuli Laine, Miika Aittala, Janne Hellsten, Jaakko Lehtinen, and
  Timo Aila.
\newblock Analyzing and improving the image quality of stylegan.
\newblock In {\em Proceedings of the IEEE/CVF conference on computer vision and
  pattern recognition}, pages 8110--8119, 2020.

\bibitem{kingma2013auto}
Diederik~P Kingma and Max Welling.
\newblock Auto-encoding variational bayes.
\newblock 2013.

\bibitem{kynkaanniemi2019improved}
Tuomas Kynk{\"a}{\"a}nniemi, Tero Karras, Samuli Laine, Jaakko Lehtinen, and
  Timo Aila.
\newblock Improved precision and recall metric for assessing generative models.
\newblock {\em Advances in Neural Information Processing Systems}, 32, 2019.

\bibitem{lee2021vitgan}
Kwonjoon Lee, Huiwen Chang, Lu Jiang, Han Zhang, Zhuowen Tu, and Ce Liu.
\newblock Vitgan: Training gans with vision transformers.
\newblock In {\em International Conference on Learning Representations}, 2021.

\bibitem{martin2021nerf}
Ricardo Martin-Brualla, Noha Radwan, Mehdi~SM Sajjadi, Jonathan~T Barron,
  Alexey Dosovitskiy, and Daniel Duckworth.
\newblock Nerf in the wild: Neural radiance fields for unconstrained photo
  collections.
\newblock In {\em Proceedings of the IEEE/CVF Conference on Computer Vision and
  Pattern Recognition}, pages 7210--7219, 2021.

\bibitem{mescheder2019occupancy}
Lars Mescheder, Michael Oechsle, Michael Niemeyer, Sebastian Nowozin, and
  Andreas Geiger.
\newblock Occupancy networks: Learning 3d reconstruction in function space.
\newblock In {\em Proceedings of the IEEE/CVF conference on computer vision and
  pattern recognition}, pages 4460--4470, 2019.

\bibitem{mildenhall2021nerf}
Ben Mildenhall, Pratul~P Srinivasan, Matthew Tancik, Jonathan~T Barron, Ravi
  Ramamoorthi, and Ren Ng.
\newblock Nerf: Representing scenes as neural radiance fields for view
  synthesis.
\newblock {\em Communications of the ACM}, 65(1):99--106, 2021.

\bibitem{miyato2018spectral}
Takeru Miyato, Toshiki Kataoka, Masanori Koyama, and Yuichi Yoshida.
\newblock Spectral normalization for generative adversarial networks.
\newblock In {\em International Conference on Learning Representations}, 2018.

\bibitem{nash2021generating}
Charlie Nash, Jacob Menick, Sander Dieleman, and Peter Battaglia.
\newblock Generating images with sparse representations.
\newblock In {\em International Conference on Machine Learning}, pages
  7958--7968. PMLR, 2021.

\bibitem{ntavelis2022arbitrary}
Evangelos Ntavelis, Mohamad Shahbazi, Iason Kastanis, Radu Timofte, Martin
  Danelljan, and Luc Van~Gool.
\newblock Arbitrary-scale image synthesis.
\newblock In {\em Proceedings of the IEEE/CVF Conference on Computer Vision and
  Pattern Recognition}, pages 11533--11542, 2022.

\bibitem{park2021nerfies}
Keunhong Park, Utkarsh Sinha, Jonathan~T Barron, Sofien Bouaziz, Dan~B Goldman,
  Steven~M Seitz, and Ricardo Martin-Brualla.
\newblock Nerfies: Deformable neural radiance fields.
\newblock In {\em Proceedings of the IEEE/CVF International Conference on
  Computer Vision}, pages 5865--5874, 2021.

\bibitem{peebles2022scalable}
William Peebles and Saining Xie.
\newblock Scalable diffusion models with transformers.
\newblock {\em arXiv preprint arXiv:2212.09748}, 2022.

\bibitem{pumarola2021d}
Albert Pumarola, Enric Corona, Gerard Pons-Moll, and Francesc Moreno-Noguer.
\newblock D-nerf: Neural radiance fields for dynamic scenes.
\newblock In {\em Proceedings of the IEEE/CVF Conference on Computer Vision and
  Pattern Recognition}, pages 10318--10327, 2021.

\bibitem{radford2015unsupervised}
Alec Radford, Luke Metz, and Soumith Chintala.
\newblock Unsupervised representation learning with deep convolutional
  generative adversarial networks.
\newblock {\em arXiv preprint arXiv:1511.06434}, 2015.

\bibitem{ramasinghe2022beyond}
Sameera Ramasinghe and Simon Lucey.
\newblock Beyond periodicity: towards a unifying framework for activations in
  coordinate-mlps.
\newblock In {\em Computer Vision--ECCV 2022: 17th European Conference, Tel
  Aviv, Israel, October 23--27, 2022, Proceedings, Part XXXIII}, pages
  142--158. Springer, 2022.

\bibitem{reed2021dynamic}
Albert~W Reed, Hyojin Kim, Rushil Anirudh, K~Aditya Mohan, Kyle Champley, Jingu
  Kang, and Suren Jayasuriya.
\newblock Dynamic ct reconstruction from limited views with implicit neural
  representations and parametric motion fields.
\newblock In {\em Proceedings of the IEEE/CVF International Conference on
  Computer Vision}, pages 2258--2268, 2021.

\bibitem{roich2022pivotal}
Daniel Roich, Ron Mokady, Amit~H Bermano, and Daniel Cohen-Or.
\newblock Pivotal tuning for latent-based editing of real images.
\newblock {\em ACM Transactions on Graphics (TOG)}, 42(1):1--13, 2022.

\bibitem{salimans2016improved}
Tim Salimans, Ian Goodfellow, Wojciech Zaremba, Vicki Cheung, Alec Radford, and
  Xi Chen.
\newblock Improved techniques for training gans.
\newblock {\em Advances in neural information processing systems}, 29, 2016.

\bibitem{sauer2021projected}
Axel Sauer, Kashyap Chitta, Jens M{\"u}ller, and Andreas Geiger.
\newblock Projected gans converge faster.
\newblock {\em Advances in Neural Information Processing Systems},
  34:17480--17492, 2021.

\bibitem{sauer2022stylegan}
Axel Sauer, Katja Schwarz, and Andreas Geiger.
\newblock Stylegan-xl: Scaling stylegan to large diverse datasets.
\newblock In {\em ACM SIGGRAPH 2022 Conference Proceedings}, pages 1--10, 2022.

\bibitem{sitzmann2020implicit}
Vincent Sitzmann, Julien Martel, Alexander Bergman, David Lindell, and Gordon
  Wetzstein.
\newblock Implicit neural representations with periodic activation functions.
\newblock {\em Advances in Neural Information Processing Systems},
  33:7462--7473, 2020.

\bibitem{sitzmann2021light}
Vincent Sitzmann, Semon Rezchikov, Bill Freeman, Josh Tenenbaum, and Fredo
  Durand.
\newblock Light field networks: Neural scene representations with
  single-evaluation rendering.
\newblock {\em Advances in Neural Information Processing Systems},
  34:19313--19325, 2021.

\bibitem{sitzmann2019scene}
Vincent Sitzmann, Michael Zollh{\"o}fer, and Gordon Wetzstein.
\newblock Scene representation networks: Continuous 3d-structure-aware neural
  scene representations.
\newblock {\em Advances in Neural Information Processing Systems}, 32, 2019.

\bibitem{skorokhodov2021adversarial}
Ivan Skorokhodov, Savva Ignatyev, and Mohamed Elhoseiny.
\newblock Adversarial generation of continuous images.
\newblock In {\em Proceedings of the IEEE/CVF Conference on Computer Vision and
  Pattern Recognition}, pages 10753--10764, 2021.

\bibitem{skorokhodov2021aligning}
Ivan Skorokhodov, Grigorii Sotnikov, and Mohamed Elhoseiny.
\newblock Aligning latent and image spaces to connect the unconnectable.
\newblock In {\em Proceedings of the IEEE/CVF International Conference on
  Computer Vision}, pages 14144--14153, 2021.

\bibitem{tan2019efficientnet}
Mingxing Tan and Quoc Le.
\newblock Efficientnet: Rethinking model scaling for convolutional neural
  networks.
\newblock In {\em International conference on machine learning}, pages
  6105--6114. PMLR, 2019.

\bibitem{tancik2020fourier}
Matthew Tancik, Pratul Srinivasan, Ben Mildenhall, Sara Fridovich-Keil, Nithin
  Raghavan, Utkarsh Singhal, Ravi Ramamoorthi, Jonathan Barron, and Ren Ng.
\newblock Fourier features let networks learn high frequency functions in low
  dimensional domains.
\newblock {\em Advances in Neural Information Processing Systems},
  33:7537--7547, 2020.

\bibitem{teague1980image}
Michael~Reed Teague.
\newblock Image analysis via the general theory of moments.
\newblock {\em Josa}, 70(8):920--930, 1980.

\bibitem{touvron2021training}
Hugo Touvron, Matthieu Cord, Matthijs Douze, Francisco Massa, Alexandre
  Sablayrolles, and Herv{\'e} J{\'e}gou.
\newblock Training data-efficient image transformers \& distillation through
  attention.
\newblock In {\em International Conference on Machine Learning}, pages
  10347--10357. PMLR, 2021.

\bibitem{tov2021designing}
Omer Tov, Yuval Alaluf, Yotam Nitzan, Or Patashnik, and Daniel Cohen-Or.
\newblock Designing an encoder for stylegan image manipulation.
\newblock {\em ACM Transactions on Graphics (TOG)}, 40(4):1--14, 2021.

\bibitem{xian2021space}
Wenqi Xian, Jia-Bin Huang, Johannes Kopf, and Changil Kim.
\newblock Space-time neural irradiance fields for free-viewpoint video.
\newblock In {\em Proceedings of the IEEE/CVF Conference on Computer Vision and
  Pattern Recognition}, pages 9421--9431, 2021.

\bibitem{yariv2020multiview}
Lior Yariv, Yoni Kasten, Dror Moran, Meirav Galun, Matan Atzmon, Basri Ronen,
  and Yaron Lipman.
\newblock Multiview neural surface reconstruction by disentangling geometry and
  appearance.
\newblock {\em Advances in Neural Information Processing Systems},
  33:2492--2502, 2020.

\bibitem{yoon2022spheresr}
Youngho Yoon, Inchul Chung, Lin Wang, and Kuk-Jin Yoon.
\newblock Spheresr: 360deg image super-resolution with arbitrary projection via
  continuous spherical image representation.
\newblock In {\em Proceedings of the IEEE/CVF Conference on Computer Vision and
  Pattern Recognition}, pages 5677--5686, 2022.

\bibitem{yu2021pixelnerf}
Alex Yu, Vickie Ye, Matthew Tancik, and Angjoo Kanazawa.
\newblock pixelnerf: Neural radiance fields from one or few images.
\newblock In {\em Proceedings of the IEEE/CVF Conference on Computer Vision and
  Pattern Recognition}, pages 4578--4587, 2021.

\bibitem{zhao2021improved}
Long Zhao, Zizhao Zhang, Ting Chen, Dimitris Metaxas, and Han Zhang.
\newblock Improved transformer for high-resolution gans.
\newblock {\em Advances in Neural Information Processing Systems},
  34:18367--18380, 2021.

\end{thebibliography}
}
\appendix

\section{Training details}
\textbf{ImageNet:} We train the Poly-INR model progressively with increasing resolution. The Poly-INR model is first trained on $200M$ images at $32\times32$ with $2048$ batch size, followed by $72M$ images at $64\times64$ with $512$ batch size, $21M$ images at $128\times128$ with $256$ batch size, $10M$ images at $256\times256$ with $128$ batch size and $2M$ images at $512\times512$ with $128$ batch size. We use learning rate of $1e^{-4}$ for the generator and $2e-{4}$ for the discriminator. We use Adam optimizer for both the generator and discriminator with $beta=(0.0, .99)$ and $eps=1e^{-8}$ and the classifier guidance loss weight is set to 8.0 starting at $128\times128$ and higher resolution. We do not use style mixing regularization and path length regularization.

\textbf{FFHQ:} We also train the Poly-INR model progressively with increasing resolution on the FFHQ dataset. We first train our model with $64\times64$ on $60M$ images using a batch size of $2048$, followed by $15M$ images at $128\times128$ with $256$ batch size and $15M$ images at $256\times256$ with $256$ batch size. The other training hyperparameters are same as the ImageNet experiments described above.

\begin{table}[h]
\centering
\caption{Poly-INR performance on FFHQ-32x32 across various levels (lvl) and model size (number of parameters in Million).}
\label{table:performance_levels}

\resizebox{0.50\textwidth}{!}{
\begin{tabular}{ccccccccc}
\toprule
\multicolumn{1}{c}{} & \multicolumn{1}{c}{\textbf{Lvl-2}}& 
\multicolumn{1}{c}{\textbf{Lvl-4}}&
\multicolumn{1}{c}{\textbf{Lvl-7}}&
\multicolumn{1}{c}{\textbf{Lvl-10}}&
\multicolumn{1}{c}{\textbf{Lvl-14}}&
\multicolumn{1}{c}{\textbf{Lvl-14}}
\\
\midrule
Feat. Dim. &512 &512&512&512&512&1024\\
Params (M) & 2.98& 5.62& 9.57 &13.52&18.79&64.74\\
FID $\downarrow$& 27.01& 3.46 & 1.92 &1.83& 1.52& 1.12\\
Precision $\uparrow$&0.85&0.68&0.67&0.68& 0.68& 0.70\\
Recall$\uparrow$&0.01&0.41&0.56&0.57&0.59&0.63\\
 \bottomrule
\end{tabular}}
\end{table}
\section{Ablation study on the number of levels and feature dimension}
We present an ablation study in Table \ref{table:performance_levels}, demonstrating the Poly-INR performance on the FFHQ-32x32 dataset as levels increase. We observe that with increasing levels, the model's performance improves. We utilize $10$ levels in our experiments because of training stability and also achieve comparable performance compared to CNN-based models. In case of training with more than $10$ levels, we can incrementally increase the number of levels by first training the model on a lower number, such as $10$, and gradually add more levels as training progresses.

In Table \ref{table:performance_levels}, we increase the model capacity either by adding more levels (layers) or increasing the feature dimension on FFHQ-32x32. We observe that when the model capacity is very small, the recall score is also very poor, but as we increase the model parameters, the recall score gets much better.

\section{Inference speeds across resolutions for our ImageNet model}
\begin{table}[h]
\centering
\caption{Inference speed (sec-per-image) of Poly-INR model trained on the ImageNet dataset across various resolutions.}
\label{table:speed_resolution}
\begin{tabular}{ccccc}
\toprule
 \multicolumn{1}{c}{\boldmath{$32^2$}}& 
\multicolumn{1}{c}{\boldmath{$64^2$}}&
\multicolumn{1}{c}{\boldmath{$128^2$}}&
\multicolumn{1}{c}{\boldmath{$256^2$}}&
\multicolumn{1}{c}{\boldmath{$512^2$}}
\\
\midrule
0.007 & 0.013 & 0.047 & 0.179 & 0.720\\
 \bottomrule
\end{tabular}
\end{table}

Table \ref{table:speed_resolution} shows the inference speed (sec-per-image) across various resolutions of the Poly-INR model trained on the ImageNet dataset on a Nvidia-RTX-6000 GPU. The Poly-INR model synthesizes each pixel independently and performs all computations at the same resolution, resulting in slower inference time at higher resolutions.

\section{Affine parameters mixing}
\begin{figure*}[]
     \centering
     \vspace{-0.2in}
         \includegraphics[width=0.84\textwidth]{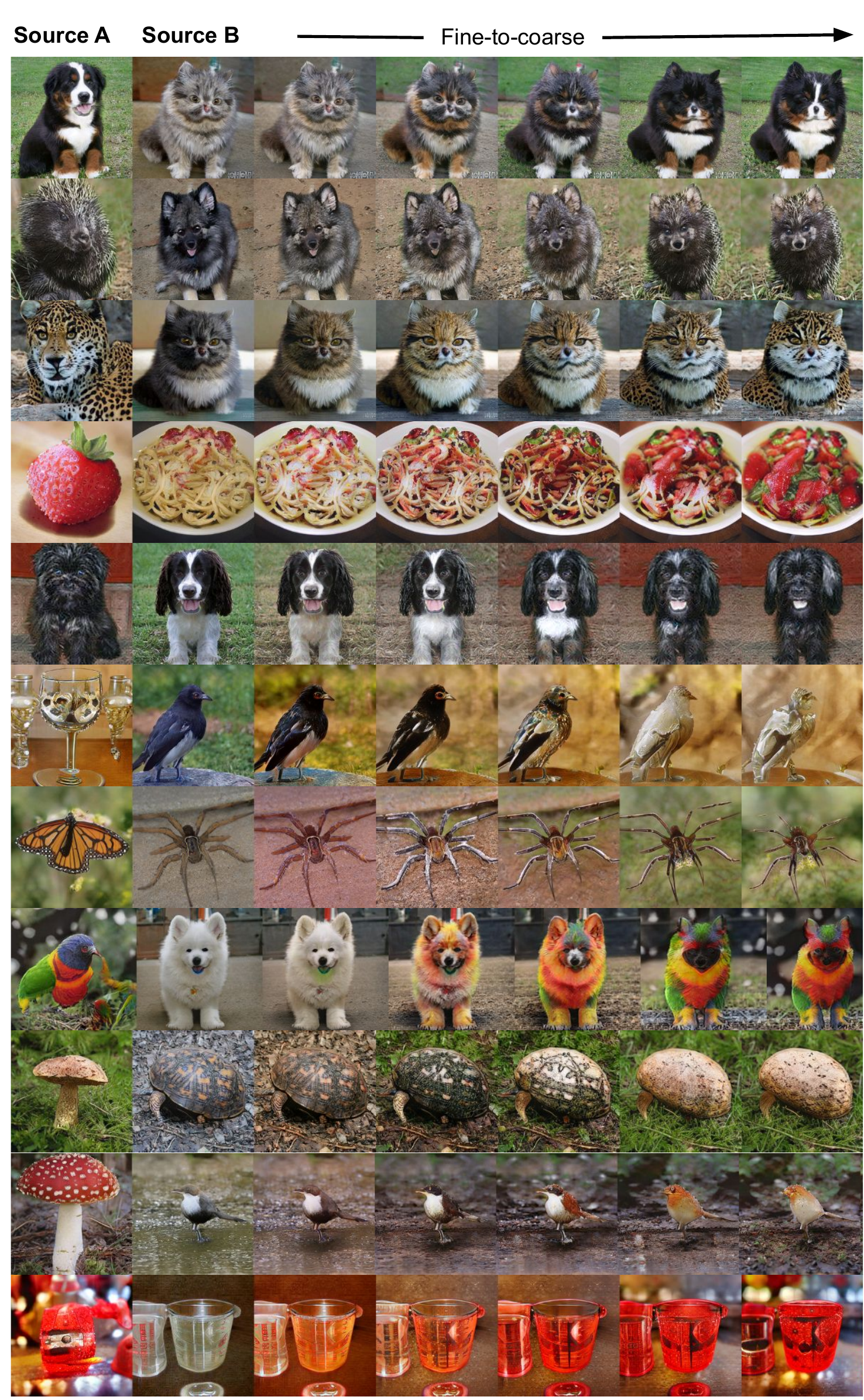}
        \vspace{-0.1in}
        \caption{Source A and B images  generated from random latent codes, and the remaining images are generated by copying the affine parameters of source A to source B at different levels. Copying the higher levels' ($8$ and $9$) affine parameters leads to finer style changes, whereas copying the middle levels' ($7$, $6$, and $5$) leads to coarse style changes. }
        \label{fig:suppli_stylemixing}
\end{figure*}

\begin{figure*}[]
     \centering
     \vspace{-0.3in}
         \includegraphics[width=0.84\textwidth]{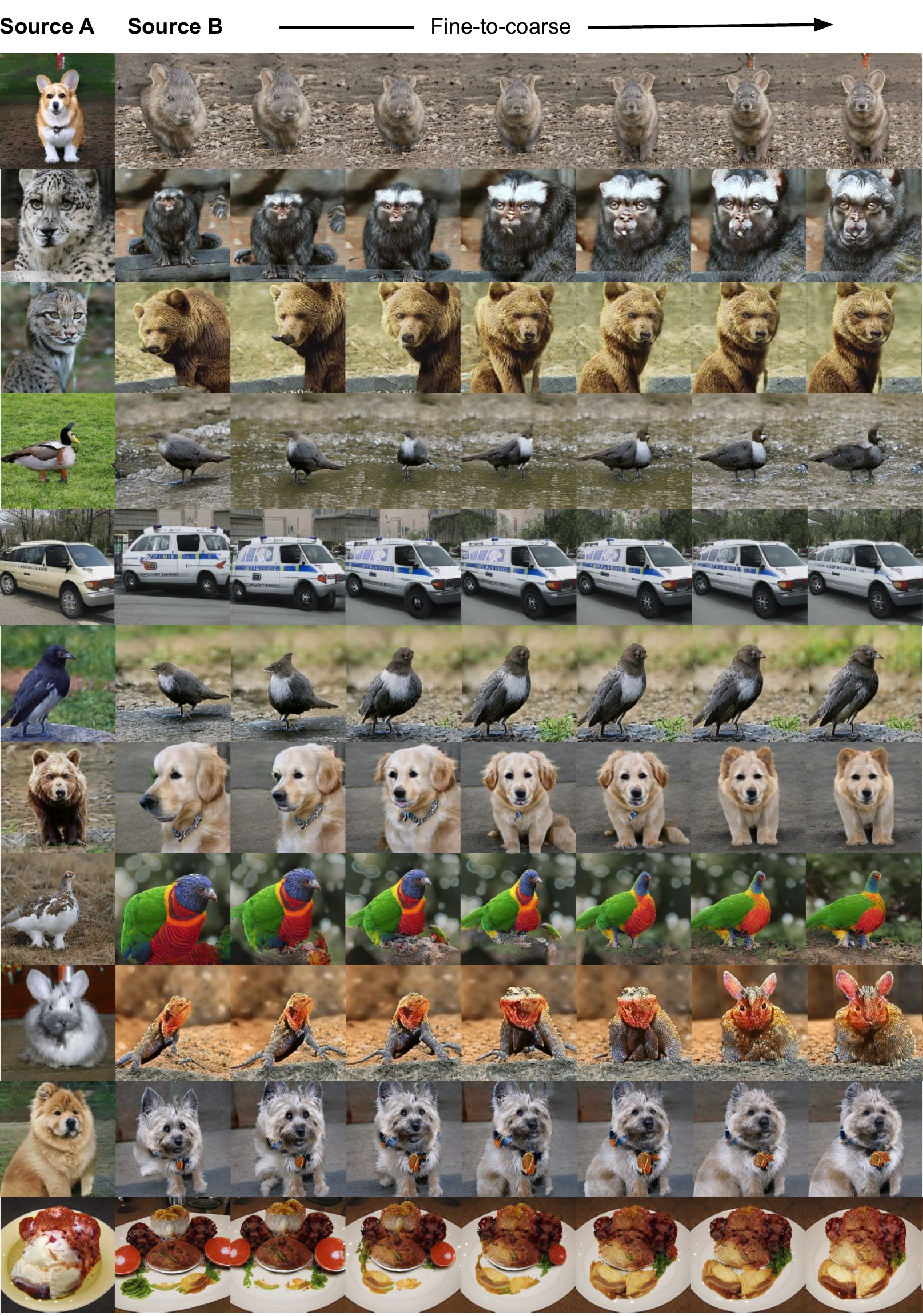}
        \vspace{-0.1in}
        \caption{Source A and B images are generated from random latent codes, and remaining images are generated by copying the affine parameters of source A to source B at different levels. Copying the initial levels' ($0$, $1$, and $2$) affine parameters leads to finer shape changes, whereas copying slightly higher levels' ($3$, $4$, and $5$) leads to coarse shape changes. }
        \label{fig:suppli_shapemixing}
\end{figure*}

An advantage of representing an image in the polynomial form is that it inherently breaks the image into shape and style. For example, the lower polynomial orders represent the object's shape, whereas the higher orders represent finer details like the style of the image. In our Poly-INR model, manipulating the lower levels' affine parameters changes the object's shape, and manipulating higher levels' affine parameters changes the style. Fig. \ref{fig:suppli_stylemixing} shows examples of style mixing from source A to source B images. In the figure, copying the affine parameters of source A to source B at higher levels ($8$ and $9$) brings fine change in the style, whereas middle levels ($5$, $6$, and $7$) bring coarse style change. Fig. \ref{fig:suppli_shapemixing} shows affine parameters mixing at initial levels ($0$-$5$). In the figure, we observe that copying the affine parameters at these levels changes the shape of the source B image to the source A image. 
\section{Interpolation}
\begin{figure*}[]
     \centering
     \vspace{-0.3in}
         \includegraphics[width=0.99\textwidth]{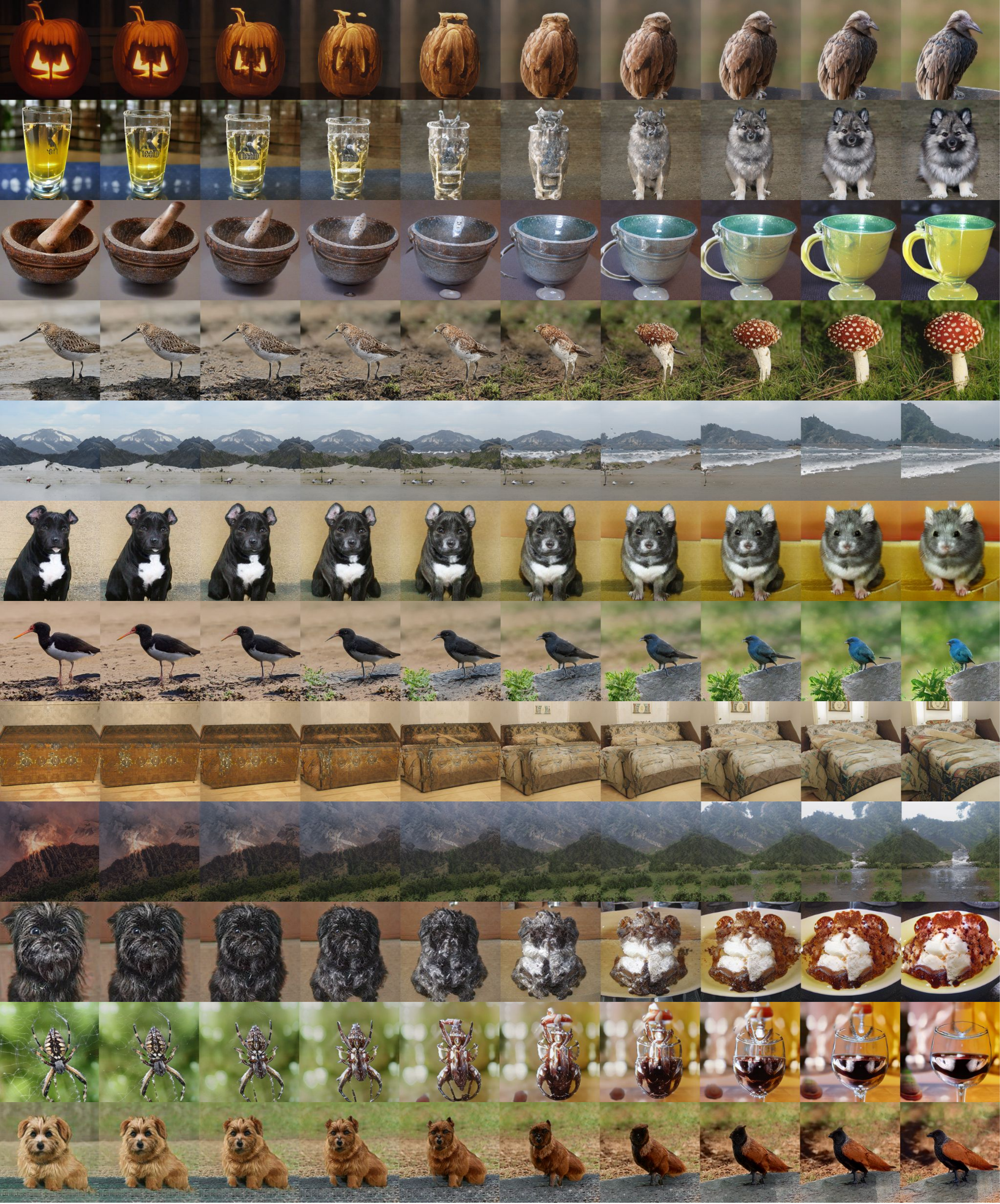}
        \vspace{-0.1in}
        \caption{The poly-INR model generates smooth interpolations between samples of different classes. }
        \label{fig:suppli_interpolation}
\end{figure*}

Fig. \ref{fig:suppli_interpolation} shows linear interpolation between samples of different classes in the affine parameters space. The Poly-INR model provides smooth interpolation between different classes. \textit{Please see the attached video for better interpolation visualization.}

\section{ Qualitative comparison with StyleGAN-XL:}
\begin{figure*}[]
     \centering
     \vspace{-0.3in}
         \includegraphics[width=0.99\textwidth]{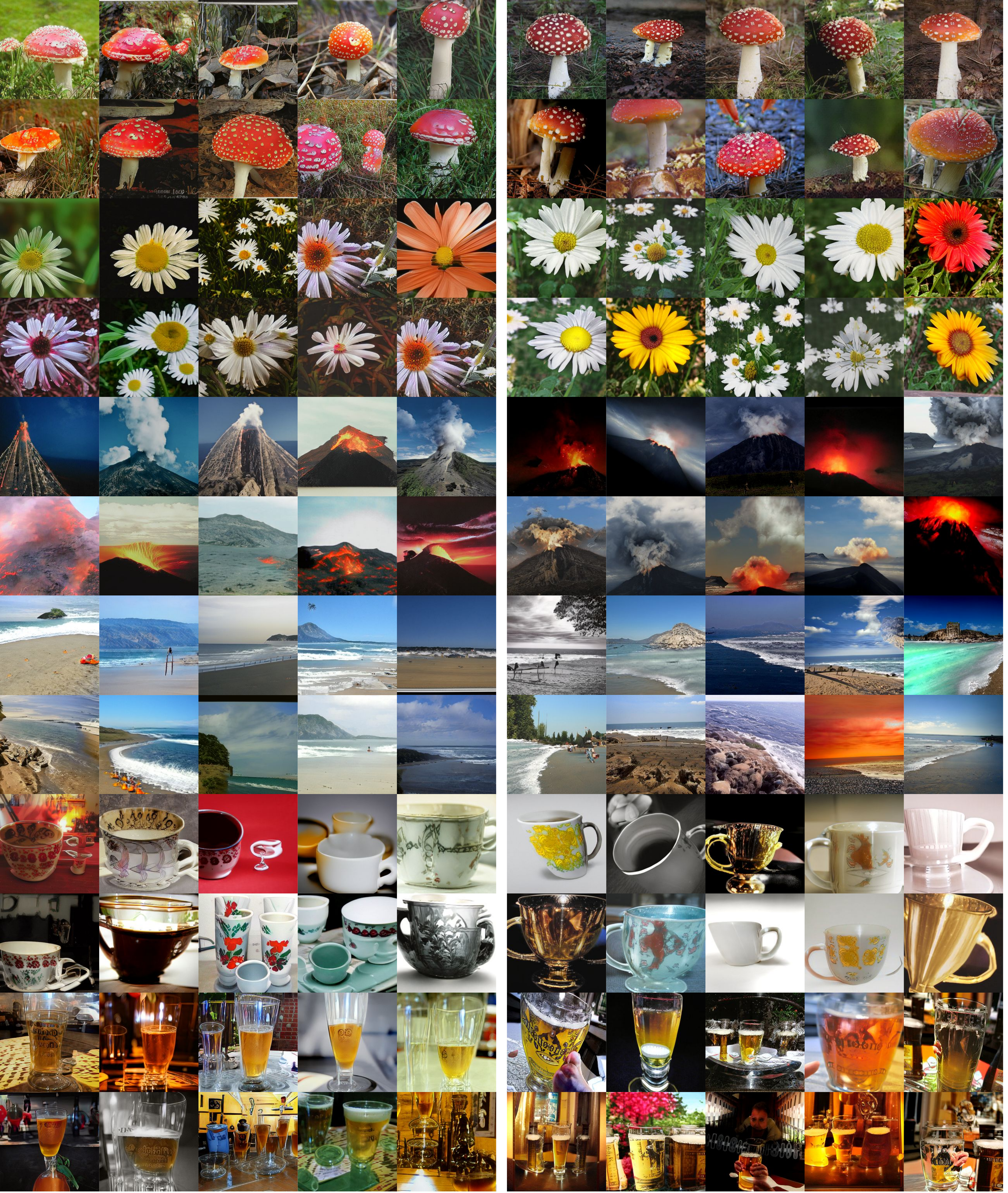}
        \vspace{-0.1in}
        \caption{Qualitative comparison between StyleGAN-XL (left column) and Poly-INR (right column). Classes from top to bottom: agaric, daisy, volcano, seashore, cup, and beer glass. }
        \label{fig:suppli_sgxl_comp1}
\end{figure*}

\begin{figure*}[]
     \centering
     \vspace{-0.3in}
         \includegraphics[width=0.99\textwidth]{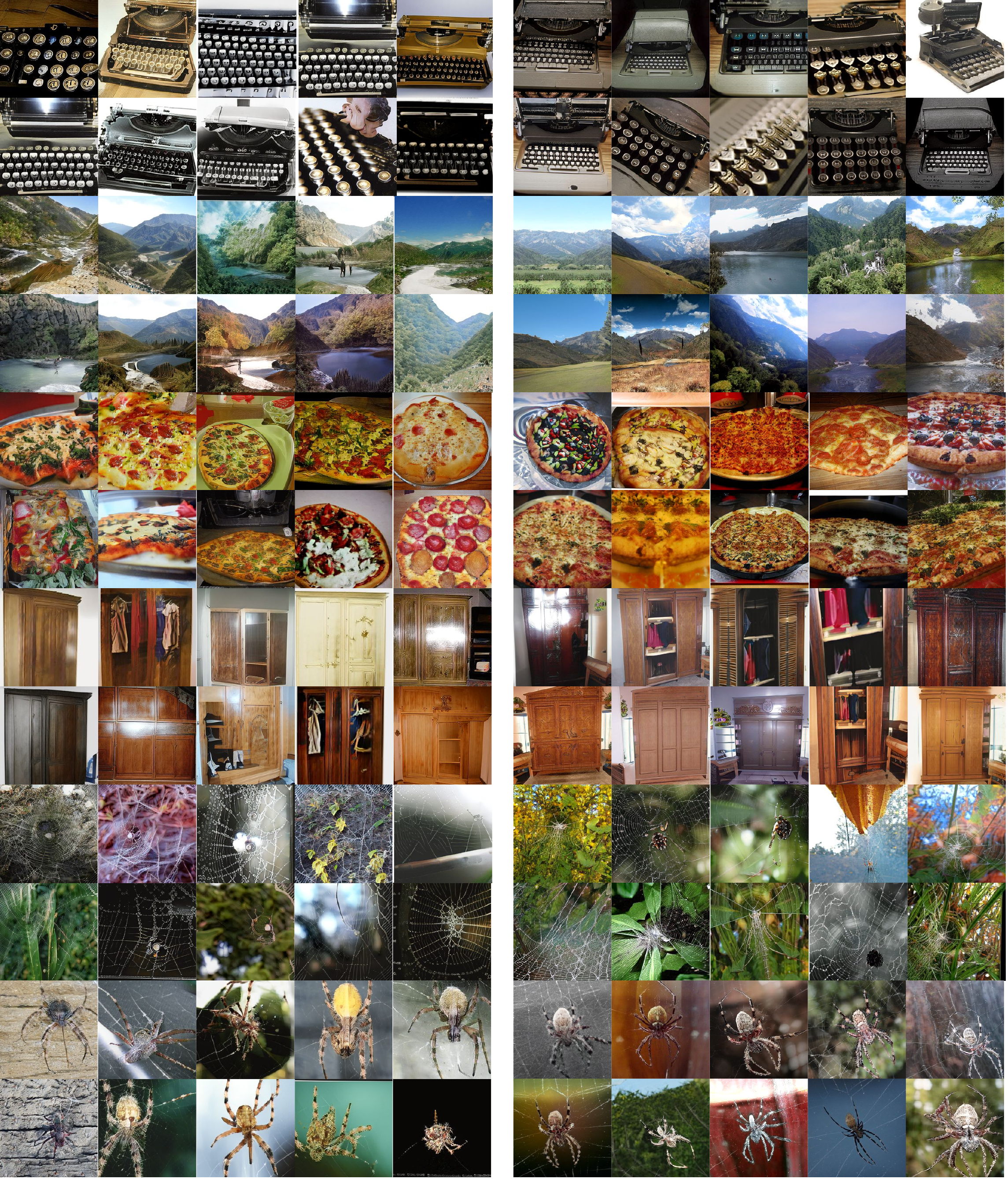}
        \vspace{-0.1in}
        \caption{Qualitative comparison between StyleGAN-XL (left column) and Poly-INR (right column). Classes from top to bottom: type writer, valley, pizza, wardrobe, spider web, barn spider. }
        \label{fig:suppli_sgxl_comp2}
\end{figure*}

\begin{figure*}[]
     \centering
     \vspace{-0.3in}
         \includegraphics[width=0.99\textwidth]{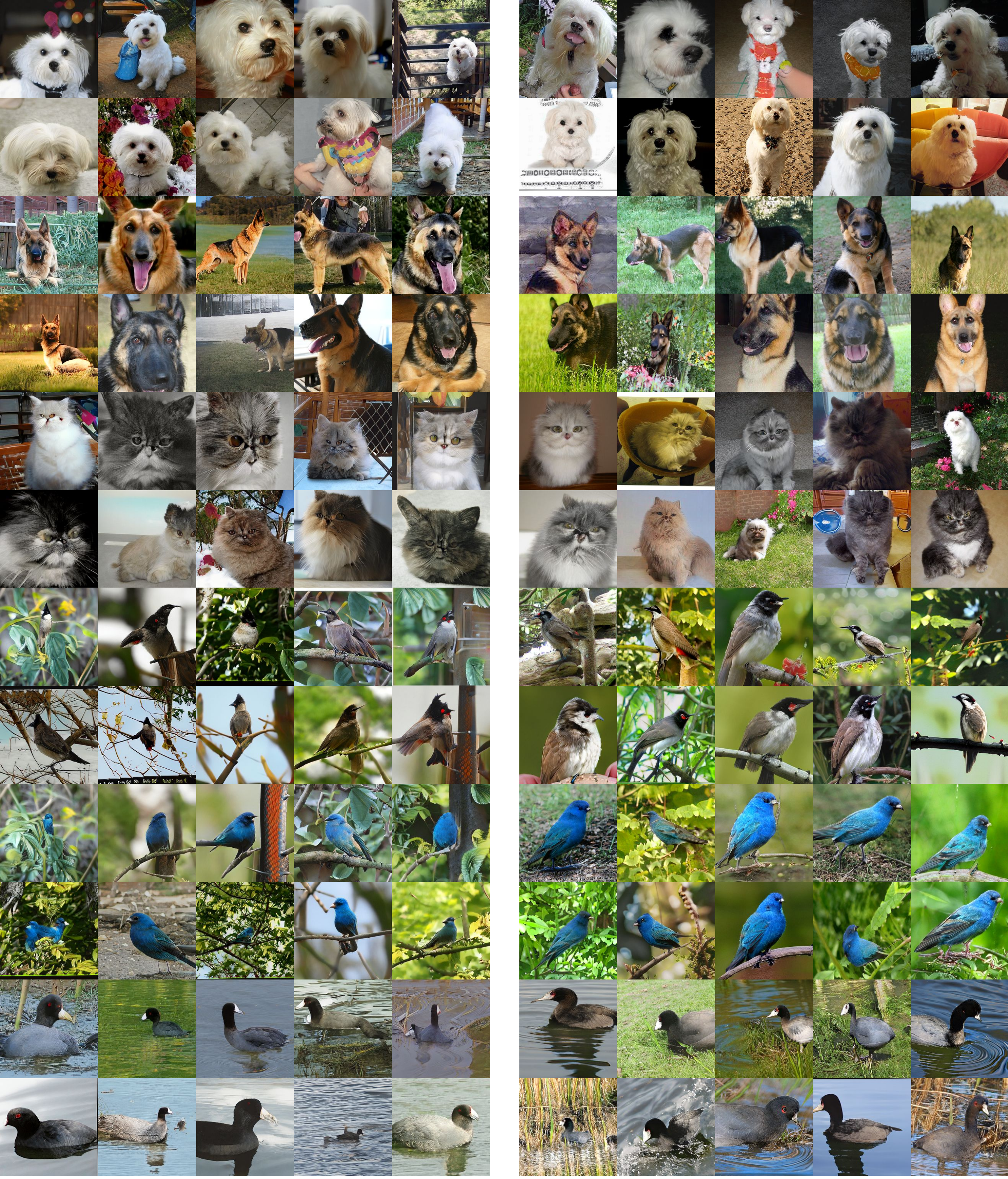}
        \vspace{-0.1in}
        \caption{Qualitative comparison between StyleGAN-XL (left column) and Poly-INR (right column). Classes from top to bottom: maltese dog, german shepherd, persian cat, bulbul, robin, american coot. }
        \label{fig:suppli_sgxl_comp3}
\end{figure*}

We also compare the quality of images generated by Poly-INR model against state-of-the-art CNN-based StyleGAN-XL model for different classes. Fig. \ref{fig:suppli_sgxl_comp1}, \ref{fig:suppli_sgxl_comp2}, and \ref{fig:suppli_sgxl_comp3} show examples of images generated from different classes for the models trained on ImageNet at $256\times256$. The Poly-INR generates samples qualitatively similar to the StyleGAN-XL model but without using any convolution or self-attention layers.

\section{Qualitative comparison with CIPS and INR-GAN on FFHQ dataset}
\begin{figure*}[]
     \centering
     \vspace{-0.3in}
         \includegraphics[width=0.99\textwidth]{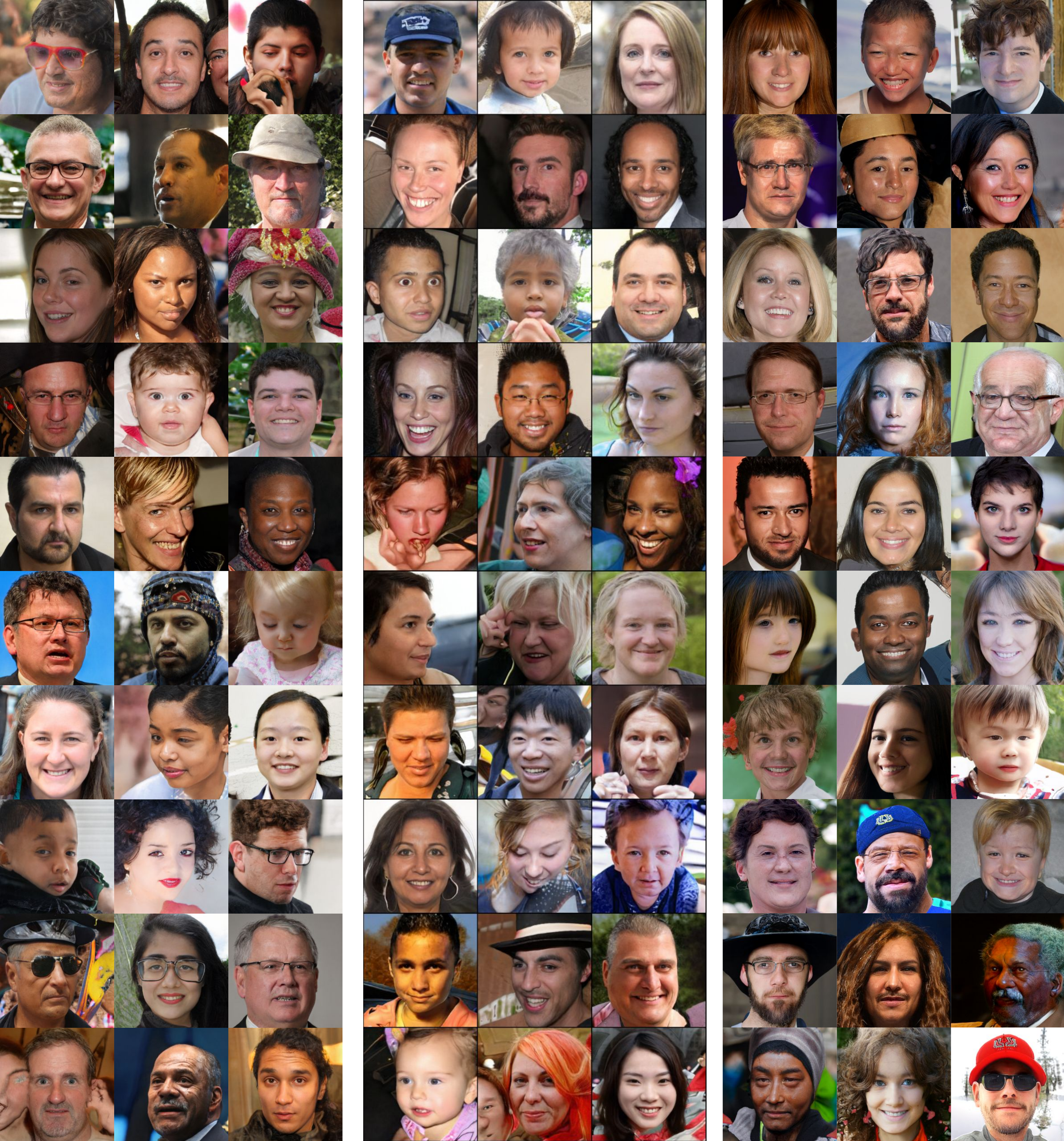}
        \vspace{-0.1in}
        \caption{Qualitative comparison between INR-GAN (left column), CIPS (middle column), and Poly-INR (right column) on FFHQ dataset at $256\times256$. }
        \label{fig:suppli_ffhq}
\end{figure*}

We also provide qualitative comparison of Poly-INR model against previously proposed INR-based generative models like CIPS and INR-GAN. Fig. \ref{fig:suppli_ffhq} shows samples generated by the three models trained on the FFHQ dataset at $256\times256$. Our Poly-INR model generates qualitatively better samples than the CIPS and INR-GAN using significantly fewer parameters.



\end{document}